\documentclass[final]{cvpr}

\usepackage{times}
\usepackage{epsfig}
\usepackage{graphicx}
\usepackage{amsmath}
\usepackage{amssymb, bm}
\usepackage{lipsum}
\usepackage[dvipsnames]{xcolor}
\usepackage{subfig}
\usepackage{stmaryrd}
\usepackage{amsthm}

\usepackage{multirow}
\usepackage{booktabs}
\usepackage{cuted}

\usepackage[pagebackref=true,breaklinks=true,colorlinks,citecolor=NavyBlue,bookmarks=false]{hyperref}
\usepackage{array}
\usepackage{color, colortbl}
\usepackage{enumitem}
\usepackage{caption}
\usepackage{multirow,bigstrut}
\usepackage{adjustbox}
\usepackage{arydshln}
\usepackage{soul}
\usepackage{nicefrac}

\definecolor{ggreen}{RGB}{15,157,88}
\definecolor{gred}{RGB}{219,68,55}

\newcommand{\R}[0]{\mathbb R}
\newcommand{\mv}[1]{\ensuremath{\bm{#1}}} 
\newcommand{\out}{\ensuremath{\bm{f}}}
\newcommand{\dis}{\ensuremath{\bm{\rho}}}
\newcommand{\wdis}{\ensuremath{\bm{\omega}}}
\newcommand{\mask}{\bm{m}}
\newcommand{\feat}{\boldsymbol{\varphi}}

\newcommand{\up}[1]{\textcolor{ggreen}{\uparrow{#1}}}  
\newcommand{\down}[1]{\textcolor{gred}{\downarrow{#1}}}

\newcolumntype{C}[1]{>{\centering\arraybackslash}p{#1}}	
\newcolumntype{L}[1]{>{\raggedright\arraybackslash}p{#1}}
\newcolumntype{R}[1]{>{\raggedleft\arraybackslash}p{#1}}
\newcolumntype{H}{>{\setbox0=\hbox\bgroup}c<{\egroup}@{}}

\newcommand{\withbox}[1]{\framebox{\scriptsize #1}}
\newcommand{\tomask}{\withbox{TO MASK}}
\newcommand{\frommask}{\withbox{FROM MASK}}

\newcommand{\stimes}{\hspace{-0.15em}\times\hspace{-0.15em}}

\newcommand{\paragr}[1]{\noindent\textbf{#1}\quad} 
\newcommand{\parag}[1]{\smallskip\noindent\textbf{#1}\quad}

\def\cvprPaperID{1179} 
\def\confYear{CVPR 2021}

\begin{document}

\title{Semantic Palette: Guiding Scene Generation with Class Proportions}
\author{Guillaume Le Moing\textsuperscript{2,}\thanks{Main part of this work was done during an internship at Valeo.ai} \hspace{0.4cm} Tuan-Hung Vu\textsuperscript{1} \hspace{0.4cm} Himalaya Jain\textsuperscript{1} \hspace{0.4cm} Patrick P\'erez\textsuperscript{1} \hspace{0.4cm} Matthieu Cord\textsuperscript{1,3}\\\\
\textsuperscript{1}Valeo.ai, Paris, France \hspace{0.8cm}  \textsuperscript{2}Inria, Paris, France\thanks{Inria, \'Ecole normale sup\'erieure, CNRS, PSL Research University} \hspace{0.8cm} 
\textsuperscript{3}Sorbonne University, Paris, France
}

\maketitle
\thispagestyle{empty}

\begin{abstract}
	Despite the recent progress of generative adversarial networks (GANs) at synthesizing photo-realistic images, producing complex urban scenes remains a challenging problem.
	Previous works break down scene generation into two consecutive phases: unconditional semantic layout synthesis and image synthesis conditioned on layouts.
	In this work, we propose to condition layout generation as well for higher semantic control: given a vector of class proportions, we generate layouts with matching composition.
	To this end, we introduce a conditional framework with novel architecture designs and learning objectives, which effectively accommodates class proportions to guide the scene generation process.
	The proposed architecture also allows partial layout editing with interesting applications.
	Thanks to the semantic control, we can produce layouts close to the real distribution, helping enhance the whole scene generation process.
	On different metrics and urban scene benchmarks, our models outperform existing baselines.
	Moreover, we demonstrate the merit of our approach for data augmentation:
	semantic segmenters trained on real layout-image pairs along with additional ones generated by our approach outperform models only trained on real pairs.
\end{abstract}


\section{Introduction}

Generative Adversarial Networks (GANs)~\cite{arjovsky2017wasserstein,goodfellow2014generative,karras2017progressive,mao2017least} have become powerful tools 
to generate photo-realistic images based on a collection of examples. When trained on real photo portraits in particular, they can produce stunning results ~\cite{karras2019style,karras2020analyzing}. However, for complex structured images like urban scenes, they still struggle to produce satisfactory results: not only do generated scenes exhibit various types of artifacts, but they are also difficult to use for downstream tasks. In automotive applications for instance, generating a wide range of synthetic driving scenes to replace or complement limited amounts of real annotated data is expected to help train better models in the future, for critical-safety tasks such as object detection and segmentation. This is not yet the case, and our work is motivated by this goal.

\begin{figure}
    \centering
    \includegraphics[width=\linewidth]{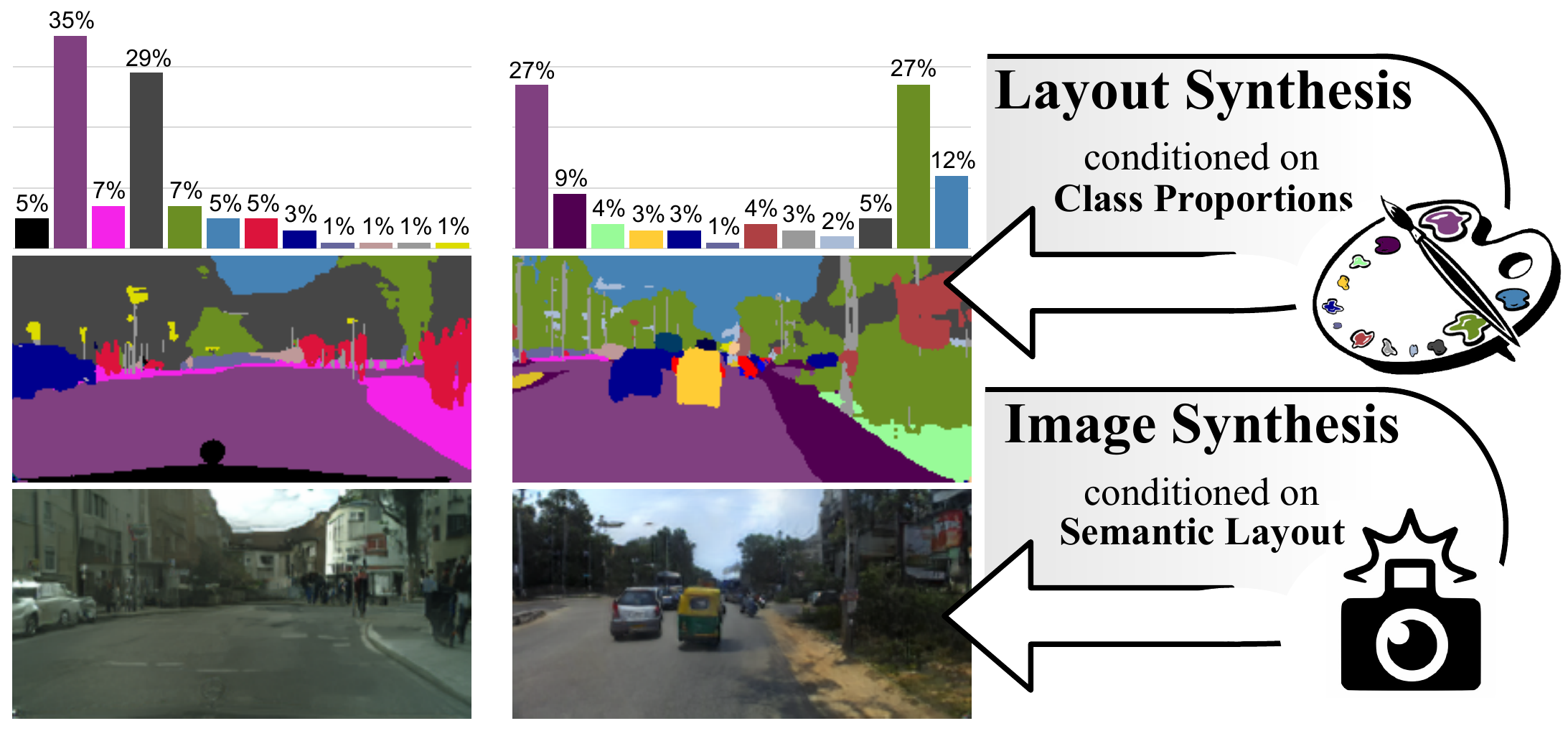}
    \vspace{-0.7cm}
    \caption{\small\textbf{Scene generation guided by semantic proportions.} The proposed approach, ``Semantic Palette'', allows a tight control of class proportions when generating semantic layouts and, conditioned on the latter, photo-realistic scenes such as urban scenes.}
    \label{fig:teaser}
    \vspace{-0.5cm}
\end{figure}

Our target application here is image semantic segmentation, the task of predicting semantic layouts, that is,  a class label (or a set of class probabilities) for each pixel of a picture. 
State-of-the-art models being fully supervised, their training requires scene examples with corresponding semantic layouts. Hence, generating this type of data with a GAN amounts to producing matching image-layout pairs. To this end, recent works advocate decoupling the synthesis process into two consecutive phases: first generating semantic layouts with plausible object arrangements~\cite{azadi2019semantic,howe2019conditional}, then translating these layouts into realistic images~\cite{park2019semantic,wang2018high}.

To improve the usability of such a pipeline, we mostly focus here on the first layout generation step.
Existing works~\cite{azadi2019semantic,howe2019conditional} cast it as a standard generative process that turns a random input code into a semantic map. While simple to use, this approach offers no real control on the modes of the output distribution~\cite{mirza2014conditional}, which is a limitation for complex scenes. In contrast, we propose to control the generation of layouts with a target distribution of semantic classes in the scene. Depending on applications, this class histogram can be manually defined, automatically derived from a true one, or sampled from a suitable distribution. To this end, we introduce a conditional layout GAN that takes a class histogram (the semantic code or \emph{palette}) as input beside the standard random noise. As a result, our full image-layout generation pipeline (Figure\,\ref{fig:teaser}) offers a simple yet powerful control over the scene composition. This ability brings benefits in various applications, ranging from real image editing to data augmentation for improved model training of a downstream task.           

Using a progressive GAN~\cite{karras2017progressive} as base architecture to generate semantic layouts, we propose novel architecture designs and learning objectives to achieve our goal.
First, we inject the semantic code throughout the progressive pipeline, \ie, at multiple intermediate scales.
To explicitly enforce the targeted class distribution while avoiding degenerate soft class assignments, we propose: a semantically-assisted activation (SAA) module along with two new learning objectives, as well as a novel residual conditional fusion module to ease the progressive propagation of the semantic target through the scales.
Lastly, we introduce a variant of the proposed framework that allows partial editing of sub-regions in existing semantic layouts.

Our main experiments are conducted on different urban scene datasets. Using suitable direct metrics, we first assess the quality of the generated layouts and of the images derived from them.  
We also assess thoroughly the merit of our approach in the light of semantic segmentation downstream task. To this end, we train segmentation models on synthesized (resp. real) data and measure their performance on real (resp. synthesized) data, as a way to compare our method with the baselines. We finally assess the ability of several approaches to improve model training through augmentation of a real-data training set.
In all experiments, Semantic Palette outperforms baselines and
produces scenes that follow better the distribution of real ones. More importantly, for real-world applications, using it to extend real datasets boosts performance in semantic segmentation. 

In summary, our main contributions are:
\begin{itemize}[noitemsep]
    \item A novel layout generative model that allows control of the distribution of semantic classes. This benefits both the quality of the images that are subsequently generated and the practical use of these images. To further enhance the quality of the generated scene-layout pairs, our method allows end-to-end training of both layout and image generators.
    \item A variant of our framework for partial editing of semantic layouts. This further benefits downstream task training and opens up interesting applications like semantic editing of real images.
    \item Extensive evaluations on three different driving benchmarks. The proposed framework significantly outperforms several baseline approaches.
\end{itemize}
\section{Related work}

\parag{Scene layout generation.}
SB-GAN~\cite{azadi2019semantic} and PGAN-CGAN~\cite{howe2019conditional} were the first GAN-based approaches proposed for this task.
The SB-GAN pipeline combines an unconditional model based on ProGAN~\cite{karras2017progressive} for generating the semantic masks and GauGAN~\cite{park2019semantic} for transforming these masks into photo-realistic images.
PGAN-CGAN~\cite{howe2019conditional} is similar but uses Pix2PixHD~\cite{wang2018high} instead as the image generator.
In the same spirit, \cite{volokitin2020decomposing} uses DCGAN~\cite{radford2015unsupervised} as the layout generator and Pix2Pix~\cite{isola2017image} as the image generator.

\parag{Conditional generative adversarial network.}
Vanilla GANs offer little control over the generation process.
In contrast, conditional GANs~\cite{mirza2014conditional} (cGANs) are designed to guide the generation with conditioning input features, that is, target attributes of the generated data.
Gradually deviating from the traditional GAN framework, alternative learning setups~\cite{odena2017conditional} and methods to better fuse the conditioning features with the generation pipeline~\cite{de2017modulating, dumoulin2016learned, liu2019learning, park2019semantic} have been proposed.
An emerging trend is to infer some generator's parameters from the conditioning input, \eg, normalization parameters~\cite{de2017modulating, dumoulin2016learned, park2019semantic}, convolutional kernels~\cite{liu2019learning}, either uniformly
~\cite{de2017modulating, dumoulin2016learned} 
or on class-specific regions
~\cite{liu2019learning, park2019semantic}, to better take into account the scene structure.

\parag{Image generation conditioned on semantic layout.}
Pix2pix, a general-purpose image-to-image translation network~\cite{isola2017image}, was among the first to produce compelling results for this conditional generation task.
Later works~\cite{liu2019learning, park2019semantic,wang2018high} have produced more realistic images at higher resolution by using a multi-scale generator, adding a feature-matching loss within the discriminator, or using instance boundary map information.
Recently, EdgeGAN~\cite{tang2020edge} proposed to generate structure and texture in parallel and blend the two together thanks to an
edge transfer module.
\section{Proposed approach}

The central goal of this work is to learn to generate plausible semantic layouts (\eg, of urban scenes), conditioned on given class proportions.
We describe our network architecture and objective functions for conditional layout generation in Sections~\ref{sec:layout_arch} and~\ref{sec:layout_objectives}.
Section~\ref{sec:img_e2e} discusses the image generation phase and how the complete pipeline with both layout and image generation can be trained end-to-end.

\subsection{Conditional layout generative network}
\label{sec:layout_arch}
\begin{figure}
    \centering
    \includegraphics[width=\linewidth]{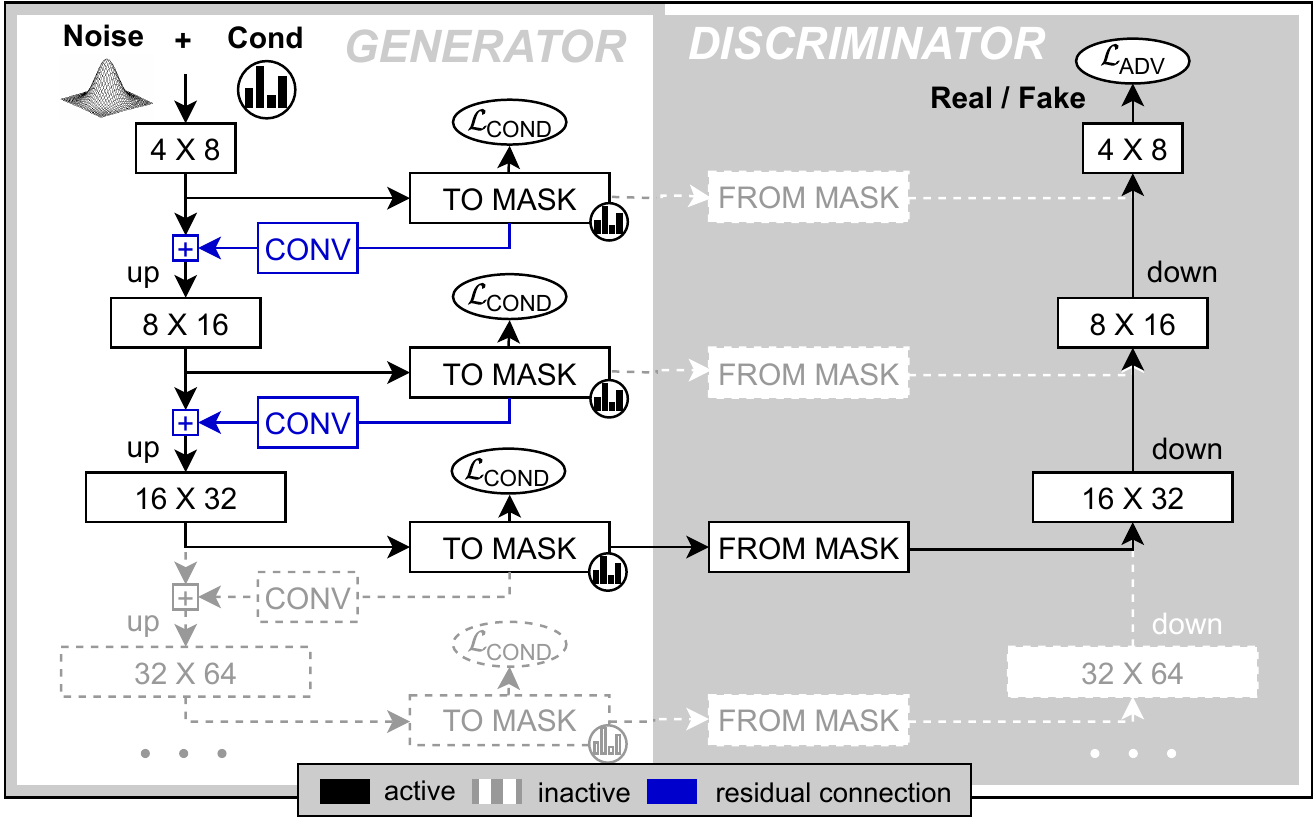}
    \vspace{-0.6cm}
    \caption{\small\textbf{Conditional synthesis of semantic layouts}. Snapshot at a certain resolution of the progressive generation (16 $\stimes$ 32 here). 
    \withbox{H $\times$ W}: two $3 \stimes 3$ convolutional layers with ReLU activations applied to feature maps of size $H\stimes W$; \tomask: turns intermediate features into soft semantic maps; \frommask: turns soft semantic maps into input features for the discriminator; \withbox{CONV}: $1 \stimes 1$ convolutional layers with ReLU activation; `` up'' and ``down'': up- and down-sampling by a factor 2. 
    }
    \label{fig:conditional_gen_arch}
    \vspace{-0.4cm}
\end{figure}

We wish to build a generative model that produces new layouts while controlling their semantic composition.
The proposed architecture builds upon ProGAN~\cite{karras2017progressive}, with new architecture designs and learning objectives to achieve our goal.
It is thus a cascade of convolutional sub-networks handling information at multiple scales, and trained in a progressive way (Figure\,\ref{fig:conditional_gen_arch}).
At each scale, intermediate features are mapped into soft semantic maps (one per class) of the corresponding resolution via the \tomask~block.
Only at the active resolution, the soft semantic layout is transformed through the \frommask~block into input features used for adversarial training.
The largest-scale output serves as the final generated outcome.

The network accepts as inputs not just a random noise vector but also a conditioning code specifying the target class distribution, \ie, the proportion of image surface that each class should occupy, for instance $50\%$ of ``sky'', $30\%$ of ``road'' and $20\%$ of ``pedestrian''.
At intermediate scales, we explicitly enforce conditioning constraints via a new \textit{semantically-assisted activation} (SAA) module operating inside \tomask~blocks.
To propagate the conditioning information from previous scales onto the next, we propose to insert a \textit{residual conditional fusion} between adjacent sub-networks.
We explain the technical details next.

\parag{Conditioning input.}
We provide our model with the necessary information for conditional generation by concatenating an input noise vector $\mv{z}$ in $\mathbb{R}^Z$ ($Z$ samples from standard Gaussian distribution) with a target normalized class histogram $\mv{t}\in\R_+^C$, with $\sum_{c=1}^{C}\mv{t}_c=1$ and $C$ the number of semantic classes.
We also use this target explicitly throughout the generation scales, as shown in the following.

\begin{figure}
	\centering
	\includegraphics[width=\linewidth]{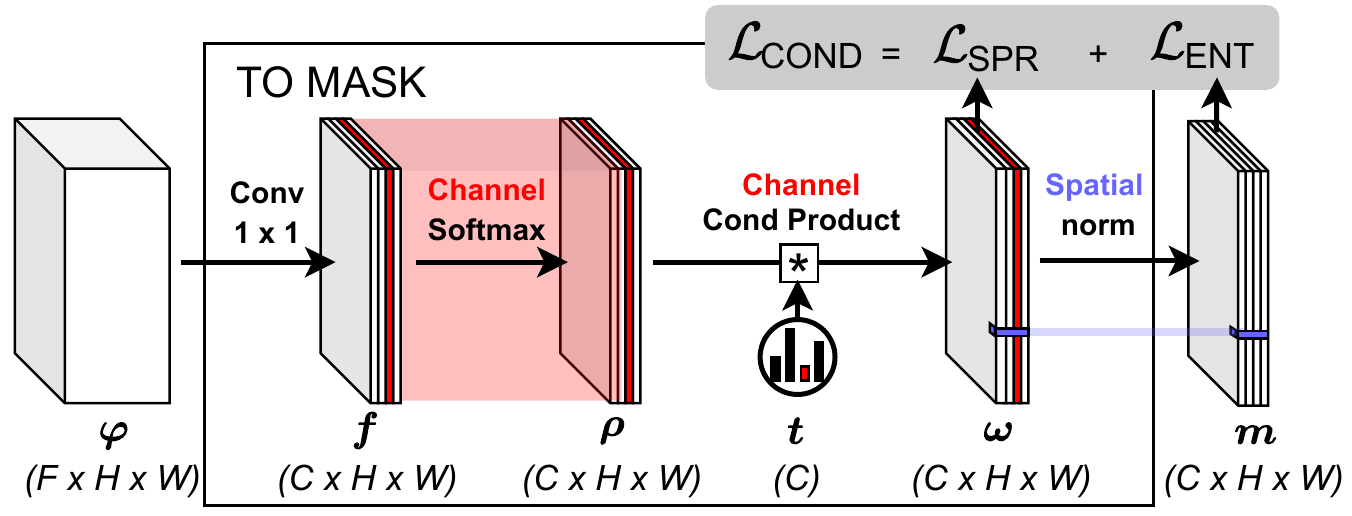}
	\vspace{-0.6cm}
	\caption{\small\textbf{Semantically-assisted activation}. 
            In \tomask~modules (Figure\,\ref{fig:conditional_gen_arch}), which produce soft semantic maps $\boldsymbol{m}$ from features $\boldsymbol{\varphi}$,
			channel softmax combined with semantic modulation forces $\wdis$ to comply with the target proportions. 
			Spread and entropy losses encourage the output mask $\boldsymbol{m}$ to retain these proportions from $\wdis$.}
			\label{fig:assisted_activation}
	\vspace{-0.4cm}
\end{figure}

\parag{Semantically-assisted activation.}
We focus here on the design of the \tomask~module, which converts deep features into soft semantic maps while respecting the prescribed class distribution.
A common choice is to use a $1\times 1$ convolutional layer to map the number of channels of the deep features to the number of semantic classes, followed by a spatial softmax activation. 
We stand out from this approach by introducing SAA, which makes again explicit use of the semantic code on top of the  generation process. This way, we expect to enforce the respect of the class proportions in the generated maps.
SAA (Figure\,\ref{fig:assisted_activation})
acts in three steps upon the $C$-channel feature maps $\out\in \mathbb{R}^{C\times H\times W}$ produced by the last convolutional layer, where $H\times W$ is the output resolution.
First, a channel-wise spatial softmax is applied to $\out$ to obtain a density map $\dis$ in $[0,1]^{C\times H\times W}$ with:
\begin{equation}
    \setlength{\abovedisplayskip}{2pt}%
    \setlength{\belowdisplayskip}{2pt}%
    \dis_{c,i,j} = \frac{\exp(\out_{c,i,j})}{\sum_{(k,\ell) \in \Omega} \exp(\out_{c,k,\ell})}\,,
    \label{eq:dis}
\end{equation}
for each class $c\in\llbracket 1, C \rrbracket$ and each pixel location $(i, j) \in \Omega = \llbracket 1, H \rrbracket \times \llbracket 1, W \rrbracket$.
Its slice $\dis_{c,:,:}$ is a normalized spatial map for class $c$. 

The next step is to use the semantic code to guide the output layout toward the target class distribution.
To this end, the channels of the density map are weighted by their corresponding target proportions to define a new map $\wdis_{c,i,j} = \mv{t}_c \cdot \dis_{c,i,j}$.
This new weighted density verifies for each class $c$, $\sum_{(i,j)\in\Omega} \wdis_{c,i,j} = \mv{t}_c$.
Thus, each class receives a ``budget'' amounting to its contribution in the semantic palette, \eg, a class with the target proportion set to zero will not be represented in the final scene.

Finally, the semantic soft map output $\mask$ in $[0,1]^{C\times H\times W}$ is obtained by $L_1$  normalization of $\wdis$, applied independently at every spatial location,
\begin{equation}
    \setlength{\abovedisplayskip}{2pt}%
    \setlength{\belowdisplayskip}{4pt}%
    \mask_{c,i,j} = \frac{\wdis_{c,i,j}}{\sum_{k \in \llbracket 1, C \rrbracket} \wdis_{k,i,j}}\, ,
    \label{eq:mask}
\end{equation}
which can be interpreted as defining at each pixel a probability distribution over all classes. As done classically, the final semantic map is obtained at each pixel by selecting the label with maximum score, \ie, $\arg\max_c \mask_{i,j,c} = \arg\max_c \wdis_{i,j,c}$. 
There is no guarantee that this hard labeling complies exactly with the target distribution, but it is tightly guided by it, as experiments will confirm.
SAA can be seen as a mechanism that transports a well-proportioned but spatially-uniform semantic map (with slice $c$ set to $\mv t_c$ everywhere) into a plausible spatial arrangement of the semantic content $\mask$.\footnote{More formally, SAA can be interpreted as the first iteration of a Sinkhorn-type algorithm~\cite{sinkhorn1967diagonal} in optimal transport, see Appendix~\ref{app:connection_to_sinkhorn}.}

In Section~\ref{sec:layout_objectives}, we will detail the two losses attached to SAA at train time, which make use of this formalization to help $\mask$ better follow the target semantic code.

\parag{Residual conditional fusion.} With the SAA design above, the output $\mask$ at an intermediate scale is already conditioned by the semantic code.
Such conditioned masks, though at lower resolutions, follow the target semantic code with realistic scene layout.
It thus makes sense to pass those signals through the generation process of higher resolutions.
This way, the \tomask~layers are still of use when moving to higher resolutions, and having access to intermediate masks produced with semantic assistance can help the network better comply with the target proportions.

To that end, we propose to include a \textit{residual conditional fusion} block before each upsampling layer (in blue in Figure\,\ref{fig:conditional_gen_arch}).
Via a $1\stimes 1$ convolutional layer, the soft mask output by the current SAA module is mapped back to features of the same size as original features $\feat$ and added to them.

\subsection{Learning objectives for layout generation}
\label{sec:layout_objectives}
We train the conditional layout synthesis network with two objectives in mind: (1) a conditional objective to help generated layouts respect the target semantic proportions and (2) an adversarial objective to ensure realism. Both could be handled simultaneously by a conditional adversarial loss~\cite{mirza2014conditional}. Here, we advocate a method that decouples the two, improving the layout quality by letting the discriminator focus on realism, as experiments will confirm.

\parag{Conditional objective.}
The conditional layout generator $G$ maps noise and target proportions pairs ($\mv z, \,\mv t$) to semantic probability masks $\mask$. To make $\mask$ follow as well as possible the target distribution, a matching loss could be used, \eg, KL-divergence between the targeted and generated class distributions.
But such a direct objective requires the non-differentiable counting of final max-score labels.
While spatial aggregation of soft class maps, $\frac{1}{HW}\sum_{i,j}\mask_{c,i,j}$, is a natural proxy for these label frequencies (coinciding with them in case of one-hot maps),  
its use can lead to undesirable solutions. For instance, a class could well be completely absent from the final layout (never being max-scoring at any location), while its soft map average matches exactly a non-zero target probability.\footnote{Details on the direct matching loss objective can be found in Appendix~\ref{app:direct_matching_loss}.} To this end, we introduce an alternative method which makes use of the proposed semantically-assisted activation.

Our novel conditional training loss has two parts: one to favor a peaky class distribution at each pixel (soft maps $\mask$ close to one-hot), 
and one to favor an even spatial semantic coverage (uniform spread of activations in $\wdis$ over pixels). Intuitively, let the semantic palette be paint of different colors in various quantities. A sufficient condition to respect proportions from the palette in the final painting is to not mix colors too much in one spot (only dominant colors will be seen) while having all the paint from the palette evenly covering the frame (no accumulation or empty spot).
The first part is translated into a loss that penalizes pixel-wise entropy of the soft masks $\mask$ generated from $(\mv z,\,\mv t)$ pairs:
\begin{equation}
    \setlength{\abovedisplayskip}{2pt}%
    \setlength{\belowdisplayskip}{4pt}%
    \mathcal{L}_{\mathrm{ENT}} = \frac{1}{HW} \mathbb{E}_{(\mv z,\,\mv t)} 
    \Big[\sum_{(i,j) \in \Omega} e_{i,j}\Big]\, ,
    \label{eq:entropy}
\end{equation}
where $e_{i,j} = - \sum_{c \in \llbracket 1, C \rrbracket}\mask_{c,i,j} \ln(\mask_{c,i,j}).$
The second part is a spread loss on the weighted density map $\wdis$. It encourages its activations to spread evenly across the image (hence to be close to  $\frac{1}{H\,W}$ at each pixel since $\sum_{c,i,j} \wdis_{c,i,j} = 1$):
\begin{equation}
    \setlength{\abovedisplayskip}{2pt}%
    \setlength{\belowdisplayskip}{4pt}%
    \mathcal{L}_{\mathrm{SPR}} = \frac{1}{H \, W} \mathbb{E}_{(\mv z,\mv t)} \Big[
    \sum_{(i,j) \in \Omega} s_{i,j} \Big]\, ,
    \label{eq:spread}
\end{equation}
where $s_{i,j} = \large( 1 - HW\sum_{c \in \llbracket 1, C \rrbracket} \wdis_{c,i,j} \large)^2\,.$
Intuitively, $\wdis$ defines a joint distribution over channels and pixel locations whose marginal over channels is defined by $\mv{t}$. The spread loss encourages the marginal over pixel locations to become uniform. This way, pixels should contribute evenly to the output semantic proportions.

The proposed conditional loss finally reads $\mathcal{L}_{\mathrm{COND}} = \mathcal{L}_{\mathrm{ENT}} + \mathcal{L}_{\mathrm{SPR}}.$ Taking advantage of the progressive structure of the generator, supervision via $\mathcal{L}_{\mathrm{COND}}$ is possible at each resolution.
The simultaneous action of SAA and of the two losses encourages the generated layouts to respect the target semantic code.
Conditional to a palette $\mv t$, if $\mathcal{L}_{\mathrm{SPR}}$ is low, in average (over $\mv z$), each spatial location receives an overall $\mv\omega$ contribution close to $(HW)^{-1}$, hence $\mv{m}\approx HW\mv{\omega}$. We get: $\mathbb{E}_{\mv z}[\sum_{i,j}\mv m_{c,i,j} | \mv t] \approx HW \mathbb{E}_{\mv z}[\sum_{i,j}\mv{\omega}_{c,i,j} | \mv t] = HW \mv t_c$ since $\sum_{i,j}\mv{\omega}_{c,i,j} = \mv t_c$ by construction. Hence, the proportions of the generated soft maps are close to the target palette on average. If $\mathcal{L}_{\mathrm{ENT}}$ is low as well, these generated soft maps will be, in addition, close to one-hot distributions at each pixel, and this average compliance with the target palette extends from the soft maps to the final layouts.

\parag{Adversarial objective.}
We train our generator to produce realistic layouts by trying to make it fool a discriminator which is jointly trained to distinguish real and generated layouts. We use the Improved WGAN loss~\cite{gulrajani2017improved} as the adversarial objective. 
Note however that real layouts are \textit{hard} masks (one-hot) while generated ones are \textit{soft} (even if  $\mathcal{L}_{\mathrm{ENT}}$ promotes, to some extent, layouts to be close to one-hot).
This discrepancy may harm the training as the generator will put lots of efforts trying to output discrete masks as well.
The solution adopted by SB-GAN~\cite{azadi2019semantic} is to apply a Gumbel-softmax~\cite{kusner2016gans} function to the generated soft masks. At each spatial location during forward pass, it samples one semantic class out of the multinomial distribution given by the semantic probabilities. During the backward pass, it behaves as a differentiable approximator of this operator.

In practice, we observed that this solution results in noisy sampled masks and using an approximator is not ideal for efficiency of training.
We propose instead to tackle this issue the other way around, \ie, by softening the real layouts.
We apply a Gaussian filter to them, with a variance adapted to the image resolution.
To ensure that the prevailing class at each pixel location remains the true one,
we use \textit{soft} semantic masks which are a weighted sum of blurred masks and original ones.
We will show the merit of our soft-layout approach compared to Gumbel-softmax in the experiments.

\subsection{Image generation and end-to-end training}
\label{sec:img_e2e}
We want to take advantage of our controllable layout generator in a complete pipeline where photo-realistic images are generated from produced layouts.
To this end, we use GauGAN~\cite{park2019semantic}, a state-of-the-art layout-to-image translation network.\footnote{{The choice of the base generative framework is orthogonal to our contributions and any improvement on it should increase the performance of our approach and the considered baselines.}}
The layout generator and the image generator are first trained individually using the default training procedure presented respectively in ProGAN~\cite{karras2017progressive} and GauGAN~\cite{park2019semantic}. Both can then be fine-tuned in an end-to-end fashion. By doing this, the layout generator benefits from additional supervision, while the image generator grows accustomed to being fed synthetic layouts which, in turn, improves the overall image quality. We use the end-to-end setup from SB-GAN~\cite{azadi2019semantic} where an additional discriminator is trained to tell apart real images from synthetic ones generated from synthetic layouts.
\section{Semantic Palette in action}

\subsection{Generating semantic codes}
\label{sec:generate_code}
SB-GAN~\cite{azadi2019semantic} directly maps noise vectors to layout-image pairs, whereas our conditional model also requires target semantic codes in inputs.
To use it for systematic data generation, we thus need a means to sample suitable semantic codes: To this end, we propose a \textit{palette generator} in the form of a Gaussian mixture model (GMM) fitted to a set of true semantic layouts, from which
the GMM 
can capture multiple meaningful modes. Vectors sampled from this GMM are then projected onto the probability $C-$simplex so as to amount to proper semantic codes. Because the exact projection is slow to compute, in practice, we simply resort to the clipping of the sampled vectors to $[0, 1]^C$ followed by $L_1$ normalization.

\subsection{Partial editing of semantic layouts}
\label{sec:partial_edit}

We extend our conditional layout generation method to layout editing, with the aim to plausibly modify an input ``real'' layout by simply manipulating its semantic palette. To this end, the generator is now conditioned on both an input layout and the target proportions. Its output is a \textit{partially edited} version of the input layout guided by the target proportions. To condition the generation on this additional input, we replace the ProGAN~\cite{karras2017progressive} convolutional blocks with the SPADE residual blocks from GauGAN \cite{park2019semantic}.

Partial editing combined with conditional generation is a powerful tool as it facilitates data augmentation with higher fidelity to real data. It also provides controlled image editing capabilities as we will see in the Experiments section.

\begin{figure}[t]
    \centering
    \includegraphics[width=\linewidth]{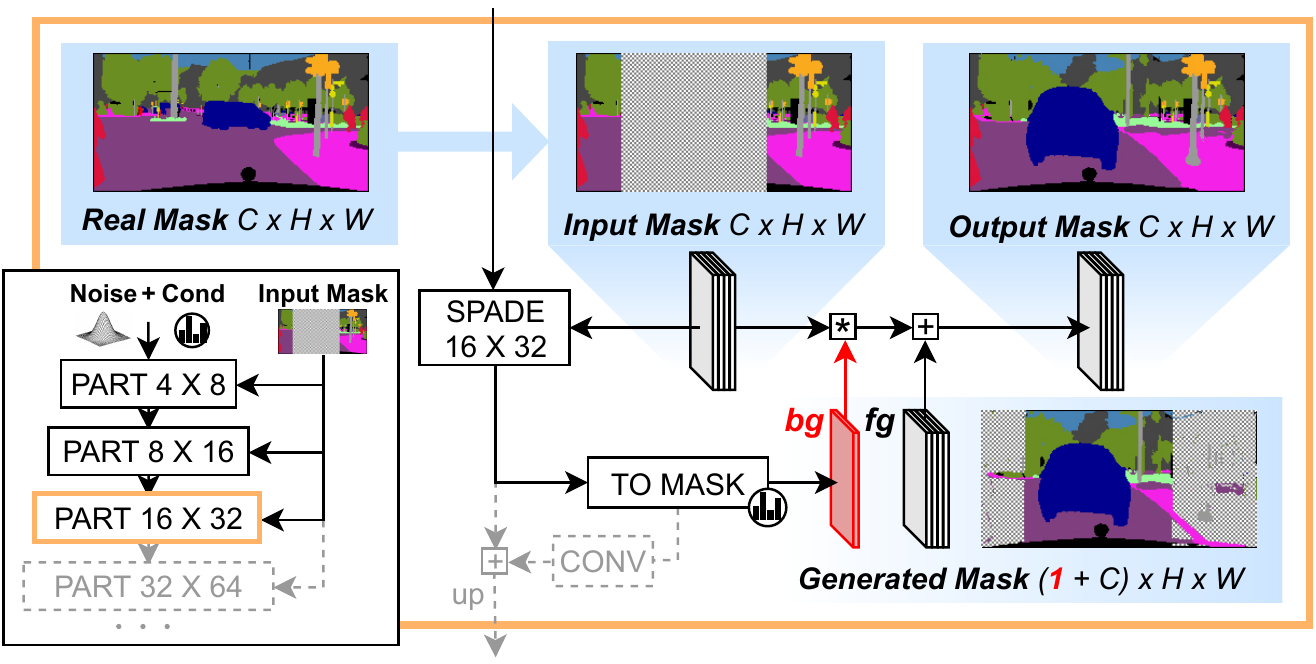}
    \vspace{-0.6cm}
    \caption{\small\textbf{Partial layout editing}. \withbox{SPADE \, H $\times$ W} replaces \withbox{H $\times$ W} from Figure\,\ref{fig:conditional_gen_arch}. It is  a SPADE~\cite{park2019semantic} residual fusion block made of two convolutional layers with conditional batchnorm and ReLU activations. The generated partial layout is merged with the input one thanks to the extra \textit{background} class (`bg').}
    \label{fig:partial_editing}
    \vspace{-0.3cm}
\end{figure}

More specifically, we consider the task of replacing an arbitrary area of the input layout.
During training, we randomly choose a rectangular patch to be replaced. The chosen patch is marked by setting all the class probabilities to $1/C$ at each pixel, while keeping the rest of the input layout as it is ($C$ is the number of semantic classes). The ``cropped" input mask is then passed to the generator through the SPADE~\cite{park2019semantic} residual blocks, see Figure\,\ref{fig:partial_editing}.
The generator synthesizes an edited version of the input mask with the ``cropped" part filled following the given target proportions for the crop. To this end, the generated mask has an additional \textit{background} class whose proportion is also set by the semantic code so as to fill the ``uncropped" part. We produce a coherent output mask by relying on this extra class to merge smoothly the generated mask to the input one. Specifically, the final output is computed as the sum of the generated mask (without background) and the input mask weighted by the background probabilities. The conditional loss $\mathcal{L}_{\mathrm{COND}}$ is applied to the generated mask while the adversarial loss is used on the output mask.
\section{Experiments}

\paragr{Datasets.} 
Evaluation is done on three urban datasets:\\
    \,\,--\,\,Cityscapes \cite{cordts2016cityscapes} is composed of 2,975 training and 500 validation scenes taken in German suburbs. All images are annotated with 33 semantic classes.\\
    \,\,--\,\,Cityscapes-25k~\cite{tao2020hierarchical} extends Cityscapes with 19,998 extra training scenes annotated by a pretrained state-of-the-art model. Note that only $19$ classes out of the $35$ original ones are effectively annotated in these additional 20K scenes.\\
    \,\,--\,\,Indian Driving Dataset (IDD) \cite{varma2019idd} contains 6,993 training and 981 validation scenes, with 35 semantic classes.

\parag{Metrics.} We use the following metrics:\\
	\,\,--\,\,Kullback–Leibler (KL) divergence between target class proportions and generated ones: It measures how well the generator respects the semantic codes.\\
	\,\,--\,\,Fréchet Segmentation Distance (FSD)~\cite{bau2019seeing}: It assesses how the overall statistics of real and synthetic layouts differ; We use real layouts from the training set.\\
	\,\,--\,\,Fréchet Inception Distance (FID)~\cite{heusel2017gans}: It is an approximate measure of generated image quality; We compute FID \wrt real images from the validation set.\\
	\,\,--\,\,GAN-test~\cite{shmelkov2018good}: We use a segmenter pretrained on real data to yield predictions for generated images. We then report the mean Intersection-over-Union both on standard official classes (mIoU) and on all classes ($\text{mIoU}^{*}$).\\
	\,\,--\,\,GAN-train~\cite{shmelkov2018good}, the opposite of GAN-test: the segmenter is trained on generated data and tested on the real validation set. Though all metrics are interesting, GAN-train better assesses the overall utility of the generated data.

\parag{Implementation details.} 
Generators are trained using ADAM~\cite{kingma2014adam}.
For segmentation, we train a DeeplabV3~\cite{chen2017rethinking} model with Stochastic Gradient Descent, $0.01$ initial $lr$, $0.9$ momentum, $5\times 10^{-4}$ weight decay, in $300$ epochs with batch-size $16$.
In all experiments, we generate layouts and images up to resolution $128\stimes 256$.
Please see Appendix~\ref{app:impl_details} for details of the palette generator.

\subsection{Conditional layout generation}
\label{sec:exp-cond}

\begin{table}
    \setlength\heavyrulewidth{0.25ex}
    \setlength{\tabcolsep}{0pt}
    \aboverulesep=0ex
    \belowrulesep=0ex
    \footnotesize
	\centering
	\begin{tabular}{@{}l@{}|@{}C{1.1cm}@{}C{1.1cm}@{}C{1.1cm}@{}C{1.1cm}@{}C{1.1cm}@{}C{1.1cm}@{}}
	    \toprule
		\multirow{2}{*}{Method} & Layout & Image & \multicolumn{2}{c}{GAN-test} & \multicolumn{2}{c}{GAN-train} \\
		\cmidrule(lr){2-2} \cmidrule(lr){3-3} \cmidrule(lr){4-5} \cmidrule(lr){6-7}
		&  KL $\downarrow$ & FID $\downarrow$ & $\text{mIoU}^{*}$ & mIoU & $\text{mIoU}^{*}$ &  mIoU \\
		\midrule
		Baseline 1\, & 1.17 & 69.2 & 33.7 & 42.8 & 29.6 & 38.5 \\
		Baseline 2\, & 0.32 & 69.0 & 35.3 & 46.9 & 30.2 & 39.4 \\
        Sem. Palette& \textbf{0.07} & {60.7} & {34.6} & {45.7} & 30.6 & 40.1 \\
        \midrule
        Sem. Palette\textsubscript{\,e2e}\, & 0.08 & \textbf{51.0} & \textbf{36.8} & \textbf{48.6} & \textbf{33.3} & \textbf{44.5} \\
        \midrule
		Oracle & - & 28.2 & - & - & 36.9 & 48.1 \\
        \bottomrule
	\end{tabular}
	\vspace{-0.2cm}
	\caption{\small\textbf{Conditional layout synthesis on Cityscapes}.
	``Oracle'': real data for FID and for training segmentator in GAN-train metric; ``e2e'': end-to-end fine-tuning; ``$\downarrow$'': smaller is better.}
	\label{tab:conditional}
	\vspace{-0.3cm}
\end{table}
\begin{figure}
	\setlength\tabcolsep{1.0pt}
	\begin{tabular}{cccc}
	    \multicolumn{4}{c}{\small(a) Interpolation between two semantic codes}\\
		\includegraphics[width=.24\linewidth]{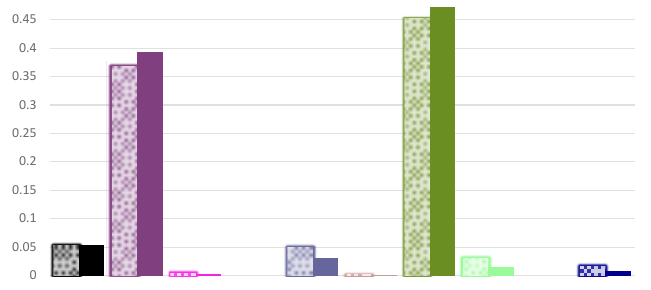} & 
		\includegraphics[width=.24\linewidth]{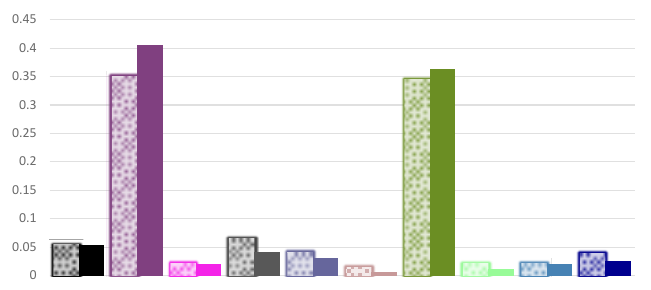} &
		\includegraphics[width=.24\linewidth]{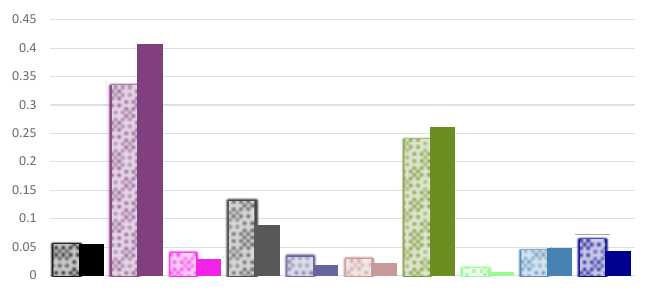} &
		\includegraphics[width=.24\linewidth]{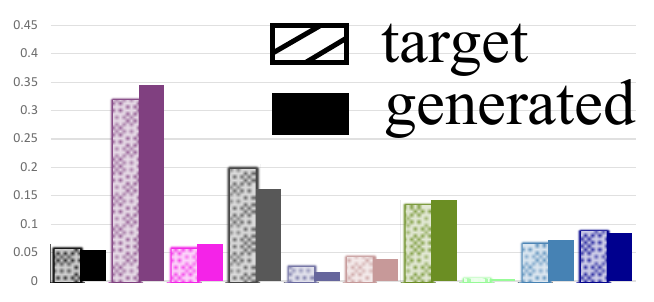} \\
		
		\includegraphics[width=.24\linewidth]{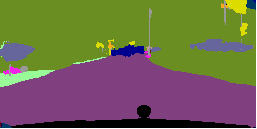} &
		\includegraphics[width=.24\linewidth]{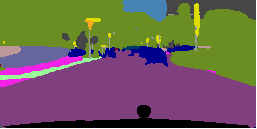} &
		\includegraphics[width=.24\linewidth]{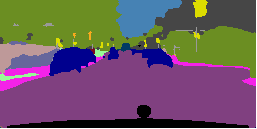} &
		\includegraphics[width=.24\linewidth]{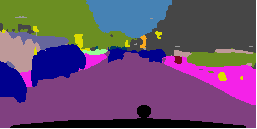} 
	\end{tabular}
	
	\begin{tabular}{cccc}
	    \multicolumn{4}{c}{\small(b) Diverse samples from one semantic code}\\
		\begin{adjustbox}{valign=t}
			\includegraphics[width=.24\linewidth]{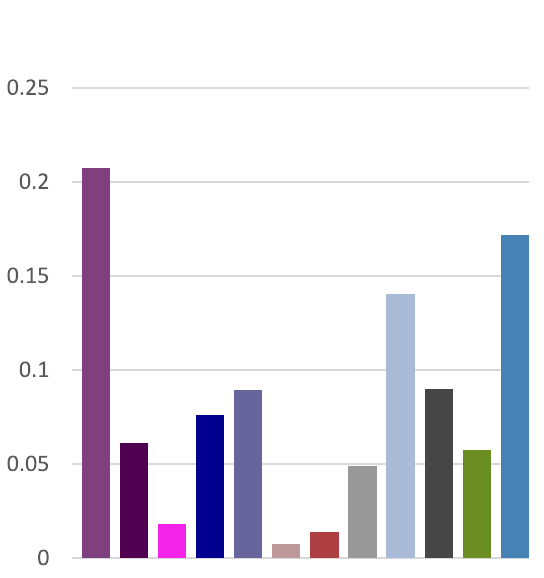}
		\end{adjustbox}
		&
		\begin{adjustbox}{valign=t}
			\begin{tabular}{@{}c@{}}
				\includegraphics[width=.24\linewidth]{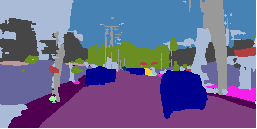}\\[-0.5mm]
				\includegraphics[width=.24\linewidth]{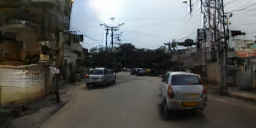}
			\end{tabular}
		\end{adjustbox}
		&
		\begin{adjustbox}{valign=t}
			\begin{tabular}{@{}c@{}}
				\includegraphics[width=.24\linewidth]{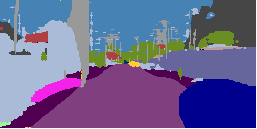}\\[-0.5mm]
				\includegraphics[width=.24\linewidth]{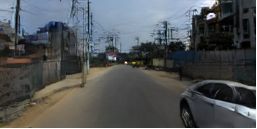}
			\end{tabular}
		\end{adjustbox}
		&
		\begin{adjustbox}{valign=t}
			\begin{tabular}{@{}c@{}}
				\includegraphics[width=.24\linewidth]{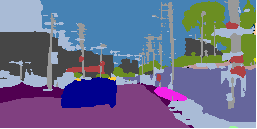}\\[-0.5mm]
				\includegraphics[width=.24\linewidth]{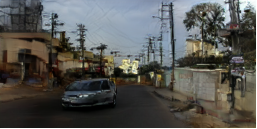}
			\end{tabular}
		\end{adjustbox}\vspace{-0.25em}\\ 
		\midrule
		\begin{adjustbox}{valign=t}
			\includegraphics[width=.24\linewidth]{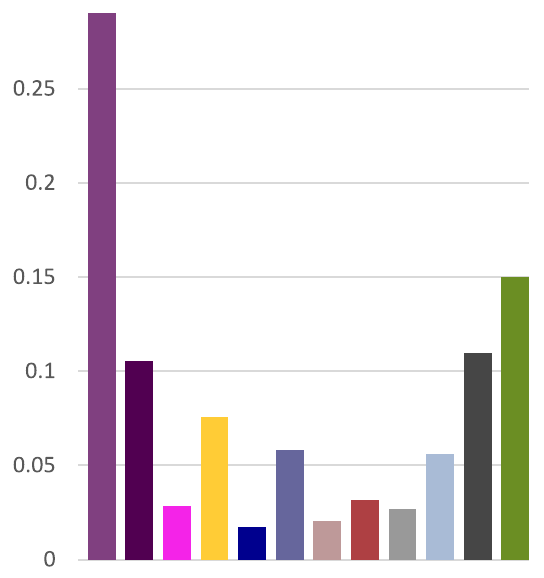}
		\end{adjustbox}
		&
		\begin{adjustbox}{valign=t}
			\begin{tabular}{@{}c@{}}
				\includegraphics[width=.24\linewidth]{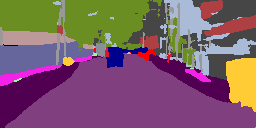}\\[-0.5mm]
				\includegraphics[width=.24\linewidth]{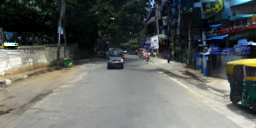}
			\end{tabular}
		\end{adjustbox}
		&
		\begin{adjustbox}{valign=t}
			\begin{tabular}{@{}c@{}}
				\includegraphics[width=.24\linewidth]{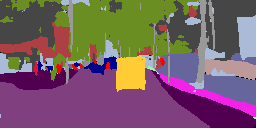}\\[-0.5mm]
				\includegraphics[width=.24\linewidth]{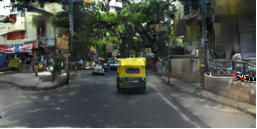}
			\end{tabular}
		\end{adjustbox}
		&
		\begin{adjustbox}{valign=t}
			\begin{tabular}{@{}c@{}}
				\includegraphics[width=.24\linewidth]{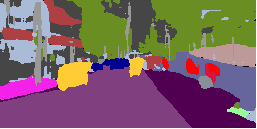}\\[-0.5mm]
				\includegraphics[width=.24\linewidth]{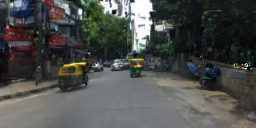}
			\end{tabular}
		\end{adjustbox}
	\end{tabular}
	
	\vspace{-0.3cm}
	\caption{\small\textbf{Conditional layout-and-scene generation.}
	(\textit{a}) For two semantic codes (left-/right-most) and interpolations between them; Class histograms in generated scenes (solid) closely follow target ones (dashed).   
	(\textit{b}) Two examples (top/bottom) of various layout-scene pairs sampled from the same semantic code (left).
	}
	\label{fig:cond_generation}
	\vspace{-0.4cm}
\end{figure}

\paragr{Comparison to conditional baselines.}
We compare Semantic Palette with two straightforward conditional layout generation baselines, \textit{baseline 1} and \textit{baseline 2}. Both accept semantic code as input.
To get layout predictions respecting target class proportions, \textit{baseline 1}  directly penalizes unsatisfying outputs via a matching loss, and \textit{baseline 2}  leverages a conditional discriminator similar to cGAN~\cite{mirza2014conditional}.
Note that the same fixed pretrained image synthesizer is used for all.

Results are reported in Table\,\ref{tab:conditional} on the Cityscapes dataset using the 4 metrics previously introduced. The direct matching method, \textit{baseline 1}, fails to reconstruct the semantic code: the KL value of $1.17$ is nearly as bad as of random guesses made on the ground-truth semantic distribution ($1.25$).
Though better on KL, \textit{baseline 2} produces images with FID comparable to \textit{baseline 1}.
Semantic Palette improves on all metrics.
Especially, we observe significant drops in KL and FID values, meaning that our conditional framework not only better respects the input semantic code, but also produces more realistic layouts.\footnote{We note that, because the same fixed image synthesizer is used in all experiments, a low FID score is a proxy indicator of layout quality.
Indeed, the image synthesizer is pretrained on real layout-scene pairs; the model is thus used to real layout inputs. The closer generated layouts are to the real distribution, the better the synthesized images are.}
While we see a slight drop compared to \textit{baseline 2} on GAN-test, more importantly, the GAN-train performance improves along with KL and FID.
The best performance is reached for Semantic Palette$_{e2e}$ after fine-tuning the layout generator and the image generator end-to-end.
By doing this, the image generator grows accustomed to synthetic layouts while the layout generator benefits from additional supervision.
Figure\,\ref{fig:cond_generation}-(a) illustrates some qualitative results.
Our layout generator clearly follows input semantic codes. 
We provide additional ablation studies on architecture choices in Section~\ref{sec:ablation}.

\begin{table*}
\setlength\heavyrulewidth{0.25ex}
\setlength{\tabcolsep}{0pt}
\aboverulesep=0ex
\belowrulesep=0ex
\footnotesize
    \centering
    \begin{tabular}{@{}l@{}|@{}C{0.85cm}@{}C{0.85cm}@{}C{0.85cm}@{}C{0.85cm}@{}C{0.85cm}@{}C{0.85cm}@{}|@{}C{0.85cm}@{}C{0.85cm}@{}C{0.85cm}@{}C{0.85cm}@{}C{0.85cm}@{}C{0.85cm}@{}|@{}C{0.85cm}@{}C{0.85cm}@{}C{0.85cm}@{}C{0.85cm}@{}C{0.85cm}@{}C{0.85cm}@{}}
    \multicolumn{1}{c}{}&\multicolumn{6}{c}{\footnotesize \rule{0pt}{2.5ex}\textbf{(a) Cityscapes}}&\multicolumn{6}{c}{\footnotesize \rule{0pt}{2.5ex}\textbf{(b) Cityscapes-25k}}&\multicolumn{6}{c}{\footnotesize \rule{0pt}{2.5ex}\textbf{(c) IDD}}\\
    \toprule
     \multirow{2}{*}{Method} & Layout & Image & \multicolumn{2}{c}{GAN-test} & \multicolumn{2}{c|}{GAN-train } & Layout & Image & \multicolumn{2}{c}{GAN-test} & \multicolumn{2}{c|}{GAN-train } & Layout & Image & \multicolumn{2}{c}{GAN-test} & \multicolumn{2}{c}{GAN-train }\\
    \cmidrule(lr){2-2} \cmidrule(lr){3-3} \cmidrule(lr){4-5} \cmidrule(lr){6-7} \cmidrule(lr){8-8} \cmidrule(lr){9-9} \cmidrule(lr){10-11} \cmidrule(lr){12-13}
    \cmidrule(lr){14-14} \cmidrule(lr){15-15} \cmidrule(lr){16-17} \cmidrule(lr){18-19}
     & FSD $\downarrow$ & FID $\downarrow$ & $\text{mIoU}^{*}$ & mIoU & $\text{mIoU}^{*}$ &  mIoU & FSD $\downarrow$ & FID $\downarrow$ & $\text{mIoU}^{*}$ & mIoU & $\text{mIoU}^{*}$ &  mIoU & FSD $\downarrow$ & FID $\downarrow$ & $\text{mIoU}^{*}$ & mIoU & $\text{mIoU}^{*}$ &  mIoU\\
    \midrule
     PCGAN~\cite{howe2019conditional} & 63.8 & 85.7 & 30.4 & 39.0 & 28.2 & 35.7 & 161.7 & 62.6 & 20.3 & 34.9 & 16.7 & 31.7 & 104.5 & 53.7 & 30.8 & 39.7 & 25.0 & 32.4 \\
     SB-GAN~\cite{azadi2019semantic} & 63.8 & 71.0 & 31.8 & 41.2 & 28.8 & 37.2 & 161.7 & 59.9 & 20.7 & 36.8 & 17.6 & 34.1 & 104.5 & 46.7 & \textbf{32.1} & \textbf{41.5} & 26.0 & 33.7\\
     Sem. Palette & \textbf{25.3} & \textbf{60.7} & \textbf{34.6} & \textbf{45.7} & \textbf{30.6} & \textbf{40.1} & \textbf{37.8} & \textbf{56.3} & \textbf{26.8} & \textbf{46.3} & \textbf{22.4} & \textbf{43.7} & \textbf{60.0} & \textbf{43.5} & 31.1 & 40.2 & \textbf{27.0} & \textbf{35.0} \\
     \midrule
     SB-GAN\textsubscript{\,e2e}~\cite{azadi2019semantic} & 20.4 & 61.8 & 34.5 & 44.7 & 29.6 & 37.0 & 148.5 & 55.1 & \textbf{28.1} & 42.9 & 24.3 & 41.8 & 116.0 & 44.8 & 31.7 & 41.0 & 27.4 & 35.6 \\
     Sem. Palette\textsubscript{\,e2e} & \textbf{11.8} & \textbf{51.0} & \textbf{36.8} & \textbf{48.6} & \textbf{33.3} & \textbf{44.5} & \textbf{61.3} & \textbf{52.8} & 27.1 & \textbf{43.9} & \textbf{24.7} & \textbf{45.1} & \textbf{40.2} & \textbf{43.2} & \textbf{32.3} & \textbf{41.7} & \textbf{27.7} & \textbf{35.9} \\
     \midrule
     Oracle & - & 28.2 & - & - & 36.9 & 48.1 & - & 30.9 & - & - & 36.5 & 53.0 & - & 26.3 & - & - & 33.8 & 43.8\\
    \bottomrule
    \end{tabular}
	\vspace{-0.3cm}
    \caption{\small\textbf{Comparison to unconditional GANs}. Same notations as in Table~\ref{tab:conditional}.}
    \label{tab:sbgan_vs_ours}
    \vspace{-0.3cm}
\end{table*}
\begin{table*}
\setlength\heavyrulewidth{0.25ex}
\setlength{\tabcolsep}{0pt}
\aboverulesep=0ex
\belowrulesep=0ex
\small
\centering
\begin{tabular}{@{}l@{}|@{}l@{}|@{}L{1.5cm}@{}L{1.5cm}@{}|@{}L{1.5cm}@{}L{1.5cm}@{}|@{}L{1.5cm}@{}L{1.5cm}@{}}
    \multicolumn{1}{c}{}&\multicolumn{1}{c}{}&\multicolumn{2}{c}{\small \rule{0pt}{2.5ex}\textbf{(a) Cityscapes}}&\multicolumn{2}{c}{\small \rule{0pt}{2.5ex}\textbf{(b) Cityscapes-25k}}&\multicolumn{2}{c}{\small \rule{0pt}{2.5ex}\textbf{(c) IDD}}\\
    \toprule
    Data & \,\, Method & \,\, $\text{mIoU}^{*}$ & mIoU & \,\, $\text{mIoU}^{*}$ & mIoU & \,\, $\text{mIoU}^{*}$ & mIoU \\
    \midrule
    Real & \,\, Baseline & \,\, 36.9 & 48.1 & \,\, 36.5 & 53.0 & \,\,  33.8 & 43.8 \\
    \midrule
    Real + Semi-Syn\,\,& \,\, GauGAN~\cite{park2019semantic} & \,\, $37.2_{\,\up{0.3}}$ & $48.2_{\,\up{0.1}}$ &\,\, $\textbf{43.0}_{\,\up{6.5}}$ & $58.0_{\,\up{5.0}}$ & \,\,  $33.6_{\,\down{0.2}}$ & $43.5_{\,\down{0.3}}$ \\
    \midrule
    \multirow{5}{*}{Real + Syn} & \,\, SB-GAN~\cite{azadi2019semantic} & \,\, $34.6_{\,\down{2.3}}$ & $45.5_{\,\down{2.6}}$ &\,\, $35.5_{\,\down{1.0}}$ & $51.4_{\,\down{1.6}}$  & \,\,  $33.5_{\,\down{0.3}}$ & $43.4_{\,\down{0.4}}$ \\
     &\,\, Sem. Palette & \,\, $38.0_{\,\up{1.1}}$ &$49.4_{\,\up{1.3}}$ &\,\, $36.9_{\,\up{0.4}}$ &$54.4_{\,\up{1.4}}$ &\,\, $33.8_{-}$ &$43.8_{-}$ \\
     &\,\, Sem. Palette (DA) & \,\, $38.6_{\,\up{1.7}}$ &$51.6_{\,\up{3.5}}$ &\,\, $38.6_{\,\up{2.1}}$ &$57.3_{\,\up{4.3}}$ &\,\, $34.5_{\,\up{0.7}}$ &$44.7_{\,\up{0.9}}$ \\
    \cmidrule(lr){2-8}
     &\,\, Sem. Palette (Part.) & \,\, $\textbf{40.7}_{\,\up{3.8}}$ &$51.9_{\,\up{3.8}}$ &\,\, ${42.4}_{\,\up{5.9}}$ &$59.1_{\,\up{6.1}}$ &\,\, $\textbf{35.6}_{\,\up{1.8}}$ &$\textbf{46.1}_{\,\up{2.3}}$ \\
     &\,\, Sem. Palette (Part. + DA) \,\, &\,\, $\textbf{40.7}_{\,\up{3.8}}$ &$\textbf{52.6}_{\,\up{4.5}}$ &\,\, ${42.5}_{\,\up{6.0}}$ &$\textbf{60.5}_{\,\up{7.5}}$ &\,\, $35.3_{\,\up{1.5}}$ &$45.8_{\,\up{2.0}}$ \\
    \bottomrule
\end{tabular}
\vspace{-0.3cm}
\caption{\small\textbf{Data-augmentation} grouped by training data regime, tested on real data. ``DA'': domain adaptation; ``Part.'': partial editing.}
\label{tab:data_augmentation}
\vspace{-0.4cm}
\end{table*}

\parag{Comparison to unconditional baselines.}
We follow~\cite{azadi2019semantic} and report performance on both Cityscapes and Cityscapes-25k datasets.
We additionally evaluate our method on the IDD dataset, to account for a very different urban landscape. 
On Cityscapes-25k, missing labels in the 20K extra images deteriorate performance of the $16$ missing classes, resulting in lower $\text{mIoU}^{*}$ as compared to Cityscapes-trained models.

We compare Semantic Palette to unconditional baselines on image-layout pairs generation (Table\,\ref{tab:sbgan_vs_ours}).
We note that SB-GAN~\cite{azadi2019semantic} does better than PGAN-CGAN~\cite{howe2019conditional} thanks to the improved image generator.
We train these baselines from scratch, nonetheless, results for SB-GAN~\cite{azadi2019semantic} on Cityscapes are in line with the ones from the original paper.

Our method outperforms the baselines in both diversity of semantic content and image quality with a clear gap in FSD, FID and GAN-train.
We facilitate the layout synthesis task by guiding explicitly the generation with semantic proportions, partially lifting the burden of figuring out the scene composition. 
We show pairs synthesized from a single target histogram on IDD in Figure\,\ref{fig:cond_generation}-(b).
Though sharing the same semantic palette, diverse scenes are produced.

\subsection{Data augmentation}
\label{sec:exp_data_augmentation}
Once trained, the image and layout generators can be used to sample new pairs, hence \textit{augmenting} the real training data. Different from standard data augmentation techniques, which only modify existing data points, synthetic models create new data points, which allows not only altering the visual appearance in the image space, but also applying structural changes in the layout space.
We consider two different data augmentation setups: (i) ``Semi-Syn'', which only relies on the pretrained image generator to synthesize images from ground-truth layouts, and (ii) ``Syn'', which uses both generators to synthesize new data pairs.
Table\,\ref{tab:data_augmentation} shows test performance of segmenters trained only on ``Real'', ``Real + Semi-Syn'', or ``'Real + Syn'' data.

\parag{Real + Semi-Syn.} A pretrained GauGAN~\cite{park2019semantic} is used as image generator.
On Cityscapes and IDD datasets, we only observe marginal changes in performance compared to the baseline.
However, when having more layouts to feed the image generator in Cityscapes-25k, the segmenter trained on augmented data significantly outperforms the baseline.
We conjecture that there is a trade-off between the quality of synthesized images and the diversity of semantic layouts: if layouts cannot provide enough diversity to counter-balance the loss of image quality, this may harm the performance.

\begin{figure}
	\setlength\tabcolsep{1.0pt}
	\tiny
	\begin{tabular}{cccc}
		(a) GT & (b) Cropped regions & (c) Generated regions & (d) Merged layout \\
		
		\includegraphics[width=.24\linewidth]{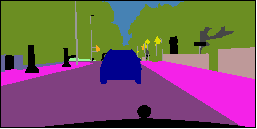} & 
		\includegraphics[width=.24\linewidth]{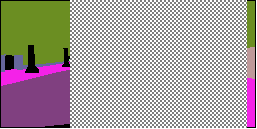} &
		\includegraphics[width=.24\linewidth]{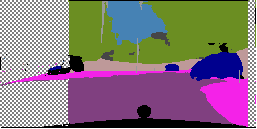} &
		\includegraphics[width=.24\linewidth]{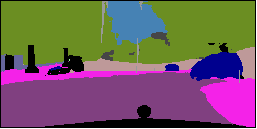} \\
		
		\includegraphics[width=.24\linewidth]{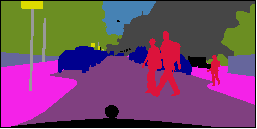} & 
		\includegraphics[width=.24\linewidth]{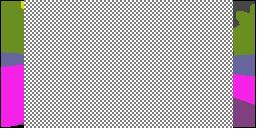} &
		\includegraphics[width=.24\linewidth]{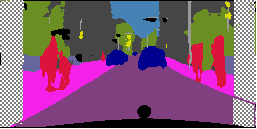} &
		\includegraphics[width=.24\linewidth]{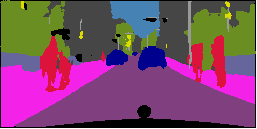}
	\end{tabular}
	\vspace{-0.3cm}
    \caption{\small\textbf{Partial editing of layouts.} The procedure consists in cropping ground-truth layouts and then synthesizing new objects within the cropped area, guided by the initial semantic proportions.}
	\label{fig:partial-edition}
	\vspace{-0.4cm}
\end{figure}

\parag{Real + Syn.} To further highlight the benefit of layout generators, we do not use the end-to-end models. The same pretrained GauGAN is used as in the ``Real + Semi-Syn'' setup.
On the three benchmarks, with {SB-GAN}~\cite{azadi2019semantic} as layout generator, we observe drops in mIoU as compared to the baseline.
The unconditional model shows its limitations in the data augmentation context where it fails to complement real data with more diverse samples, resulting in negative results.
In contrast, Semantic Palette consistently improves upon baselines, except for IDD dataset where the performance is unchanged.
These results demonstrate the merits of our pipeline for data augmentation.

We propose several variants of our method to further push its performance.
First, to alleviate distribution gaps between synthetic and real data, we adopt AdvEnt~\cite{vu2019advent}, a domain adaptation technique for semantic segmentation.\footnote{See Appendix~\ref{app:impl_details} for implementation details.}
This strategy is used to ensure synthetic and real supervisions are consistent.
This domain adaptation (DA) technique boosts further the performance of our approach (Table\,\ref{tab:data_augmentation}). 
Second, we test a variant using the partial layout editing method presented in Section~\ref{sec:partial_edit} and illustrated in Figure\,\ref{fig:partial-edition}.
The ensuing performance (Table\,\ref{tab:data_augmentation}) demonstrates the clear benefit of leveraging extra real information in the generation process, \ie, partial areas and semantic proportions.
A straightforward combination of the two proposed strategies achieves the best performance on two benchmarks.

In Appendix~\ref{app:comparison_standard_aug}, we provide further results and discuss the additional use of standard data augmentation during training.

\subsection{Ablation studies}
\label{sec:ablation}
We report the results of an ablation study in Table\,\ref{tbl:ablation}.
Using residual fusion significantly decreases KL, FSD, and FID values, highlighting the benefit of leveraging lower-scale information.
We then achieve further improvements in KL and FID with multi-scale training.
Using a strategy based on soft ground-truth masks, as detailed in Section~\ref{sec:layout_objectives}, instead of Gumbel-softmax improves KL score by a large margin while preserving comparable FID scores.
Our final model using the palette generator presented in Section~\ref{sec:generate_code} achieves the best FID, with a slightly worse KL score compared to the model using ground-truth semantic codes.

\begin{table}
    \setlength\heavyrulewidth{0.25ex}
    \setlength{\tabcolsep}{0pt}
    \aboverulesep=0ex
    \belowrulesep=0ex
	\footnotesize
	\centering
	\begin{tabular}{@{}C{1.23cm}@{}C{1.23cm}@{}C{1.23cm}@{}C{1.23cm}@{}|@{}C{1.1cm}@{}C{1.1cm}@{}C{1.1cm}@{}}
	    \toprule
		\multirow{2}{*}{\begin{tabular}[c]{@{}c@{}}Residual\\ Fusion\end{tabular}}& \multirow{2}{*}{\begin{tabular}[c]{@{}c@{}}Multi\\ Scale\end{tabular}} & \multirow{2}{*}{\begin{tabular}[c]{@{}c@{}}Soft GT\\ Masks\end{tabular}} & \multirow{2}{*}{\begin{tabular}[c]{@{}c@{}}Palette\\ Gen.  \end{tabular}} & \multicolumn{2}{c}{Layout} & Image \\
		\cmidrule(lr){5-6} \cmidrule(lr){7-7}
		& & & &  KL $\downarrow$ & FSD $\downarrow$ & FID $\downarrow$ \\
		\midrule
		& & & & 0.32 & 33.9 & 70.6 \\
        \checkmark & & & & 0.13 & \textbf{23.4} & 65.5 \\
        \checkmark & \checkmark & & & 0.11 & 37.1 & 63.3\\
        \checkmark & \checkmark & \checkmark & & \textbf{0.03} & 24.1 & 64.3 \\
        \checkmark & \checkmark & \checkmark & \checkmark & 0.07 & 25.3 & \textbf{60.7} \\
        \bottomrule
	\end{tabular}
	\vspace{-0.3cm}
	\caption{\small\textbf{Semantic Palette ablation on Cityscapes.} First row: model with only SAA; In models with ``Soft GT Masks''  unmarked, Gumbel-softmax is used; If ``Palette Gen.'' is unmarked, ground-truth codes are used instead of generated ones.}
	\label{tbl:ablation}
	\vspace{-0.4cm}
\end{table}

\subsection{Face editing}

\begin{figure}
	\setlength\tabcolsep{1.0pt}
	\tiny
	\begin{tabular}{cccccc}
	    \multicolumn{6}{c}{\small (a) Hair manipulation.}\\[1mm]
		\includegraphics[width=.158\linewidth]{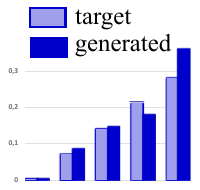} &
		\includegraphics[width=.158\linewidth]{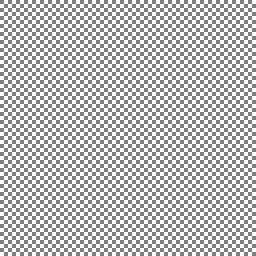} & 
		\includegraphics[width=.158\linewidth]{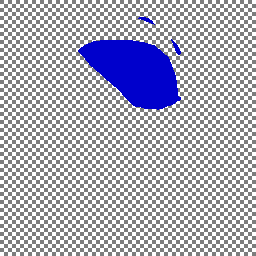} & 
		\includegraphics[width=.158\linewidth]{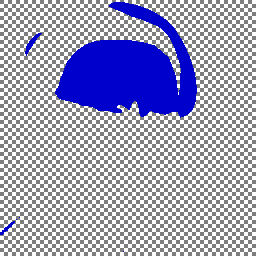} & 
		\includegraphics[width=.158\linewidth]{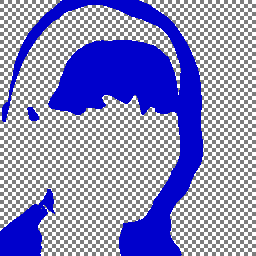} & 
		\includegraphics[width=.158\linewidth]{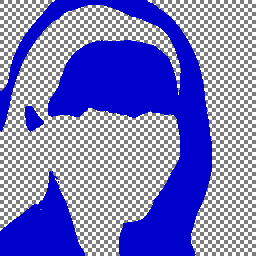} \\[-0.5mm]
		
		\includegraphics[width=.158\linewidth]{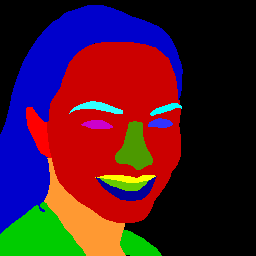} &
		\includegraphics[width=.158\linewidth]{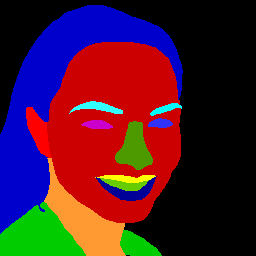} & 
		\includegraphics[width=.158\linewidth]{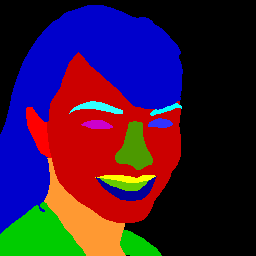} & 
		\includegraphics[width=.158\linewidth]{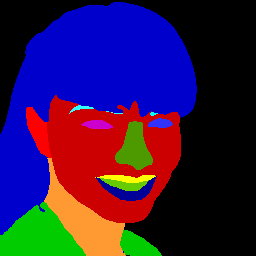} & 
		\includegraphics[width=.158\linewidth]{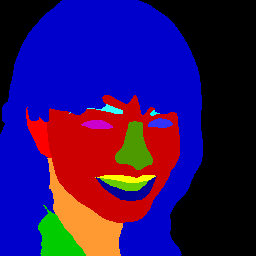} & 
		\includegraphics[width=.158\linewidth]{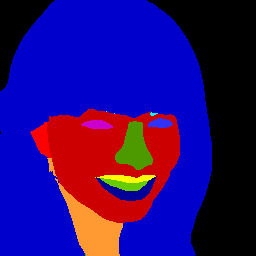} \\[-0.5mm]
		
		\includegraphics[width=.158\linewidth]{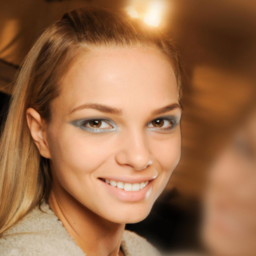} &
		\includegraphics[width=.158\linewidth]{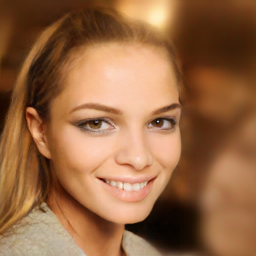} &
		\includegraphics[width=.158\linewidth]{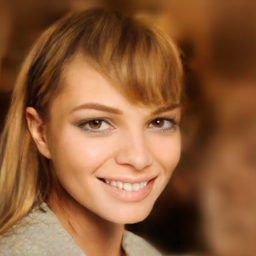} &
		\includegraphics[width=.158\linewidth]{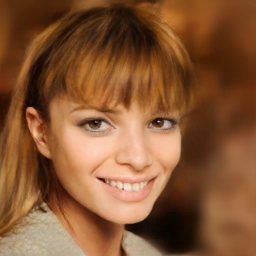} &
		\includegraphics[width=.158\linewidth]{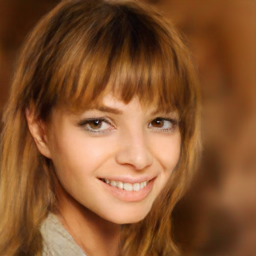} &
		\includegraphics[width=.158\linewidth]{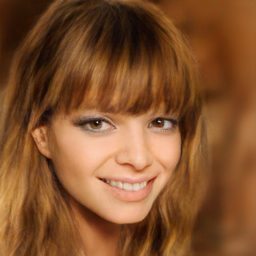} \\ [-0.5mm]
		\multicolumn{1}{c!\vrule}{Ground-truth} & \multicolumn{5}{c}{Interpolation to target proportions} \\[-0.5mm]
		\arrayrulecolor{black!50}\midrule
	\end{tabular}
		
	\begin{tabular}{cccccc}
		\multicolumn{6}{c}{\small (b) Diverse semantic attributes} manipulation.\\[1mm]
		\includegraphics[width=.158\linewidth]{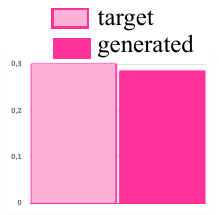} &
		\includegraphics[width=.158\linewidth]{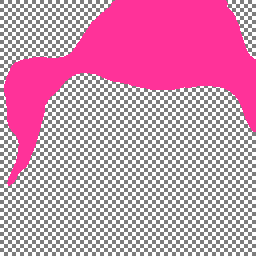} & 
		\includegraphics[width=.158\linewidth]{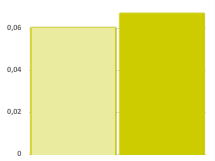} & 
		\includegraphics[width=.158\linewidth]{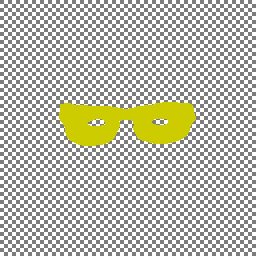} & 
		\includegraphics[width=.158\linewidth]{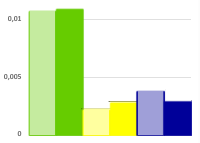} & 
		\includegraphics[width=.158\linewidth]{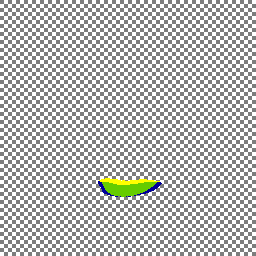} \\[-0.5mm]
		
		\includegraphics[width=.158\linewidth]{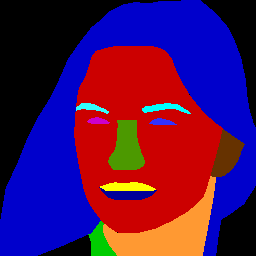} &
		\includegraphics[width=.158\linewidth]{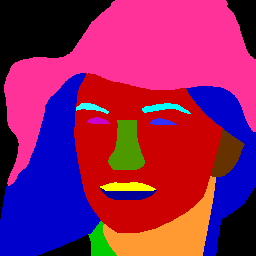} & 
		\includegraphics[width=.158\linewidth]{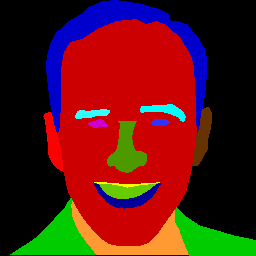} & 
		\includegraphics[width=.158\linewidth]{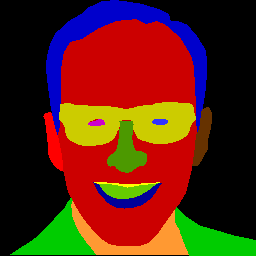} & 
		\includegraphics[width=.158\linewidth]{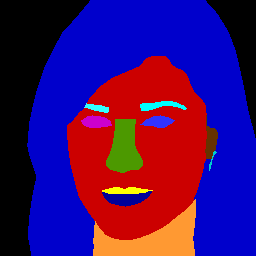} & 
		\includegraphics[width=.158\linewidth]{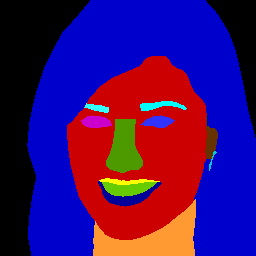} \\[-0.5mm]
		
		\includegraphics[width=.158\linewidth]{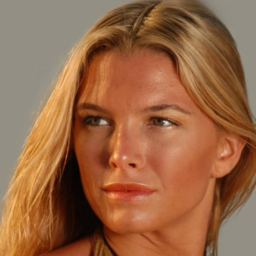} &
		\includegraphics[width=.158\linewidth]{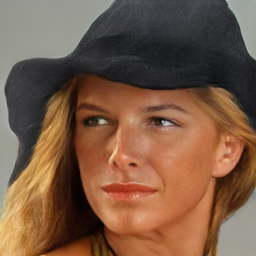} &
		\includegraphics[width=.158\linewidth]{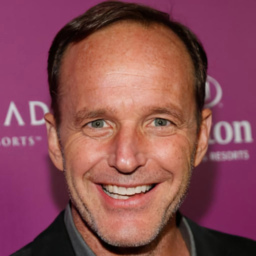} &
		\includegraphics[width=.158\linewidth]{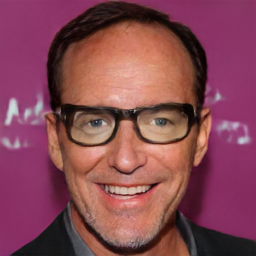} &
		\includegraphics[width=.158\linewidth]{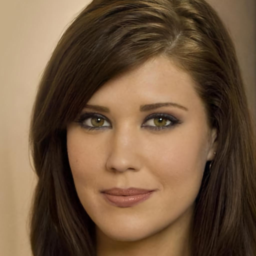} &
		\includegraphics[width=.158\linewidth]{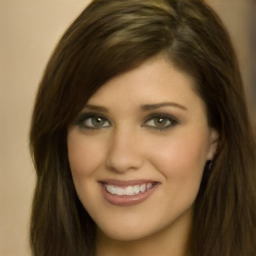} \\[-0.5mm]
		Ground-truth & \multicolumn{1}{c!\vrule}{Generated} & Ground-truth & \multicolumn{1}{c!\vrule}{Generated} & Ground-truth & Generated\\[-0.5mm]
		\arrayrulecolor{black!50}\midrule
	\end{tabular}
	\vspace{-0.3cm}
	\caption{\small \textbf{Application of Semantic Palette to face editing} at resolution $256\stimes 256$. In (a), we illustrate the fine-controlled editing of layouts by gradually increasing the budget for the hair. Edits are convincing both in the layout and image spaces. Thanks to the novelty loss, there is little overlap between original and additional hair. In (b), we show the editing of diverse semantic attributes. Although we have a unique layout generator, we can perform very different edits. Moreover, one can play with latent codes to generate various edits for the same proportion of semantic attributes.}
	\label{fig:face_manipulation}
	\vspace{-0.4cm}
\end{figure}

We showcase the new editing capabilities offered by the combination of conditional and partial layout generation on face images, using the CelebAMask-HQ dataset~\cite{karras2017progressive, lee2020maskgan, liu2015deep, dapo2020}. For the image synthesis, we use a pretrained SEAN~\cite{zhu2020sean} model, an upgrade of GauGAN~\cite{park2019semantic} where one can fix independently the style of individual semantic classes. We use it to maintain the content while editing the semantic structure. Our method, illustrated in Figure\,\ref{fig:face_manipulation}, allows one to adjust semantic attributes by a chosen amount with realistic details. This is achieved by simply modifying class proportions, avoiding the tedious task of direct manual editing of the original face layout. For this task, we do not crop the original layouts but allocate some budget for semantic additions. When target content is already present in the original layout, the generator will be inclined to replicate the original content as it fully satisfies both conditional and adversarial objectives; \eg, to increase the amount of hair, it will copy the existing hair as long as the proportion matches since it is the definition of realism for the discriminator.
To counter this undesired behaviour, we introduce a novelty loss that encourages edits to be different from original semantic classes (details in Appendix~\ref{app:novelty_loss}). 

\section{Conclusion}

We have proposed the Semantic Palette, a new framework for scene generation, and editing, guided by semantic proportions. Using novel architecture designs and learning objectives -- semantically assisted activation and residual conditional fusion coupled with novel conditional losses --, it generates plausible scene layouts with class proportions close to target ones, which then translate into realistic images. Experiments assess the superior quality of the generated layout-image pairs as well as their utility for downstream-task training: used in particular to augment an original real-data set, they deliver performance gain in semantic segmentation.

~

\paragr{Acknowledgements.} \small{Part of this work was done using HPC resources from GENCI–IDRIS (Grant 2020-AD011012227). We would like to thank Jean Ponce for useful comments.}

\clearpage
{
\fontsize{9.6pt}{9.6pt}\selectfont
\bibliographystyle{ieee_fullname}
\bibliography{paper-arxiv}

\begin{thebibliography}{10}\itemsep=-1pt

\bibitem{arjovsky2017wasserstein}
Martin Arjovsky, Soumith Chintala, and L{\'e}on Bottou.
\newblock {Wasserstein GAN}.
\newblock In {\em ICML}, 2017.

\bibitem{azadi2019semantic}
Samaneh Azadi, Michael Tschannen, Eric Tzeng, Sylvain Gelly, Trevor Darrell,
  and Mario Lucic.
\newblock Semantic bottleneck scene generation.
\newblock {\em arXiv preprint}, 2019.

\bibitem{bau2019seeing}
David Bau, Jun-Yan Zhu, Jonas Wulff, William Peebles, Hendrik Strobelt, Bolei
  Zhou, and Antonio Torralba.
\newblock Seeing what a gan cannot generate.
\newblock In {\em ICCV}, 2019.

\bibitem{chen2017rethinking}
Liang-Chieh Chen, George Papandreou, Florian Schroff, and Hartwig Adam.
\newblock Rethinking atrous convolution for semantic image segmentation.
\newblock {\em arXiv preprint}, 2017.

\bibitem{cordts2016cityscapes}
Marius Cordts, Mohamed Omran, Sebastian Ramos, Timo Rehfeld, Markus Enzweiler,
  Rodrigo Benenson, Uwe Franke, Stefan Roth, and Bernt Schiele.
\newblock The {Cityscapes} dataset for semantic urban scene understanding.
\newblock In {\em CVPR}, 2016.

\bibitem{cuturi2013sinkhorn}
Marco Cuturi.
\newblock Sinkhorn distances: Lightspeed computation of optimal transport.
\newblock In {\em NeurIPS}, 2013.

\bibitem{dapo2020}
Arnaud Dapogny, Matthieu Cord, and Patrick Perez.
\newblock The missing data encoder: Cross-channel image completion with
  hide-and-seek adversarial network.
\newblock In {\em AAAI}, 2020.

\bibitem{de2017modulating}
Harm De~Vries, Florian Strub, J{\'e}r{\'e}mie Mary, Hugo Larochelle, Olivier
  Pietquin, and Aaron~C Courville.
\newblock Modulating early visual processing by language.
\newblock In {\em NeurIPS}, 2017.

\bibitem{dumoulin2016learned}
Vincent Dumoulin, Jonathon Shlens, and Manjunath Kudlur.
\newblock A learned representation for artistic style.
\newblock In {\em ICLR}, 2017.

\bibitem{goodfellow2014generative}
Ian Goodfellow, Jean Pouget-Abadie, Mehdi Mirza, Bing Xu, David Warde-Farley,
  Sherjil Ozair, Aaron Courville, and Yoshua Bengio.
\newblock Generative adversarial nets.
\newblock In {\em NeurIPS}, 2014.

\bibitem{gulrajani2017improved}
Ishaan Gulrajani, Faruk Ahmed, Martin Arjovsky, Vincent Dumoulin, and Aaron~C
  Courville.
\newblock Improved training of {Wasserstein GANs}.
\newblock In {\em NeurIPS}, 2017.

\bibitem{heusel2017gans}
Martin Heusel, Hubert Ramsauer, Thomas Unterthiner, Bernhard Nessler, and Sepp
  Hochreiter.
\newblock {GANs} trained by a two time-scale update rule converge to a local
  {Nash} equilibrium.
\newblock In {\em NeurIPS}, 2017.

\bibitem{howe2019conditional}
Jonathan Howe, Kyle Pula, and Aaron~A Reite.
\newblock Conditional generative adversarial networks for data augmentation and
  adaptation in remotely sensed imagery.
\newblock In {\em Applications of Machine Learning}, 2019.

\bibitem{isola2017image}
Phillip Isola, Jun-Yan Zhu, Tinghui Zhou, and Alexei~A Efros.
\newblock Image-to-image translation with conditional adversarial networks.
\newblock In {\em CVPR}, 2017.

\bibitem{karras2017progressive}
Tero Karras, Timo Aila, Samuli Laine, and Jaakko Lehtinen.
\newblock Progressive growing of {GANs} for improved quality, stability, and
  variation.
\newblock In {\em ICLR}, 2018.

\bibitem{karras2019style}
Tero Karras, Samuli Laine, and Timo Aila.
\newblock A style-based generator architecture for generative adversarial
  networks.
\newblock In {\em CVPR}, 2019.

\bibitem{karras2020analyzing}
Tero Karras, Samuli Laine, Miika Aittala, Janne Hellsten, Jaakko Lehtinen, and
  Timo Aila.
\newblock Analyzing and improving the image quality of {styleGAN}.
\newblock In {\em CVPR}, 2020.

\bibitem{kingma2014adam}
Diederik~P Kingma and Jimmy Ba.
\newblock Adam: A method for stochastic optimization.
\newblock In {\em ICLR}, 2015.

\bibitem{kusner2016gans}
Matt~J Kusner and Jos{\'e}~Miguel Hern{\'a}ndez-Lobato.
\newblock {GANs} for sequences of discrete elements with the gumbel-softmax
  distribution.
\newblock {\em arXiv preprint}, 2016.

\bibitem{lee2020maskgan}
Cheng-Han Lee, Ziwei Liu, Lingyun Wu, and Ping Luo.
\newblock {MaskGan}: Towards diverse and interactive facial image manipulation.
\newblock In {\em CVPR}, 2020.

\bibitem{liu2019learning}
Xihui Liu, Guojun Yin, Jing Shao, Xiaogang Wang, and Hongsheng Li.
\newblock Learning to predict layout-to-image conditional convolutions for
  semantic image synthesis.
\newblock In {\em NeurIPS}, 2019.

\bibitem{liu2015deep}
Ziwei Liu, Ping Luo, Xiaogang Wang, and Xiaoou Tang.
\newblock Deep learning face attributes in the wild.
\newblock In {\em ICCV}, 2015.

\bibitem{mao2017least}
Xudong Mao, Qing Li, Haoran Xie, Raymond~YK Lau, Zhen Wang, and Stephen
  Paul~Smolley.
\newblock Least squares generative adversarial networks.
\newblock In {\em ICCV}, 2017.

\bibitem{mirza2014conditional}
Mehdi Mirza and Simon Osindero.
\newblock Conditional generative adversarial nets.
\newblock {\em arXiv preprint}, 2014.

\bibitem{odena2017conditional}
Augustus Odena, Christopher Olah, and Jonathon Shlens.
\newblock Conditional image synthesis with auxiliary classifier gans.
\newblock In {\em ICML}, 2017.

\bibitem{park2019semantic}
Taesung Park, Ming-Yu Liu, Ting-Chun Wang, and Jun-Yan Zhu.
\newblock Semantic image synthesis with spatially-adaptive normalization.
\newblock In {\em CVPR}, 2019.

\bibitem{radford2015unsupervised}
Alec Radford, Luke Metz, and Soumith Chintala.
\newblock Unsupervised representation learning with deep convolutional
  generative adversarial networks.
\newblock In {\em ICLR}, 2016.

\bibitem{shmelkov2018good}
Konstantin Shmelkov, Cordelia Schmid, and Karteek Alahari.
\newblock How good is my {GAN}?
\newblock In {\em ECCV}, 2018.

\bibitem{sinkhorn1967diagonal}
Richard Sinkhorn.
\newblock Diagonal equivalence to matrices with prescribed row and column sums.
\newblock {\em The American Mathematical Monthly}, 1967.

\bibitem{tang2020edge}
Hao Tang, Xiaojuan Qi, Dan Xu, Philip~HS Torr, and Nicu Sebe.
\newblock Edge guided {GANs} with semantic preserving for semantic image
  synthesis.
\newblock {\em arXiv preprint}, 2020.

\bibitem{tao2020hierarchical}
Andrew Tao, Karan Sapra, and Bryan Catanzaro.
\newblock Hierarchical multi-scale attention for semantic segmentation.
\newblock {\em arXiv preprint}, 2020.

\bibitem{varma2019idd}
Girish Varma, Anbumani Subramanian, Anoop Namboodiri, Manmohan Chandraker, and
  CV Jawahar.
\newblock {IDD}: A dataset for exploring problems of autonomous navigation in
  unconstrained environments.
\newblock In {\em WACV}, 2019.

\bibitem{volokitin2020decomposing}
Anna Volokitin, Ender Konukoglu, and Luc Van~Gool.
\newblock Decomposing image generation into layout prediction and conditional
  synthesis.
\newblock In {\em CVPRw}, 2020.

\bibitem{vu2019advent}
Tuan-Hung Vu, Himalaya Jain, Maxime Bucher, Matthieu Cord, and Patrick
  P{\'e}rez.
\newblock {AdvEnt}: Adversarial entropy minimization for domain adaptation in
  semantic segmentation.
\newblock In {\em CVPR}, 2019.

\bibitem{wang2018high}
Ting-Chun Wang, Ming-Yu Liu, Jun-Yan Zhu, Andrew Tao, Jan Kautz, and Bryan
  Catanzaro.
\newblock High-resolution image synthesis and semantic manipulation with
  conditional {GANs}.
\newblock In {\em CVPR}, 2018.

\bibitem{zhou2017ade}
Bolei Zhou, Hang Zhao, Xavier Puig, Sanja Fidler, Adela Barriuso, and Antonio
  Torralba.
\newblock Scene parsing through {ADE}20k dataset.
\newblock In {\em CVPR}, 2017.

\bibitem{zhu2020sean}
Peihao Zhu, Rameen Abdal, Yipeng Qin, and Peter Wonka.
\newblock {SEAN}: Image synthesis with semantic region-adaptive normalization.
\newblock In {\em CVPR}, 2020.

\end{thebibliography}
}

\clearpage
\normalsize
\appendix
\section{Connection to Sinkhorn algorithm}
\label{app:connection_to_sinkhorn}

To carry on the discussion initiated in Section~\ref{sec:layout_arch}, we here elaborate on the connection between our SAA module and the Sinkhorn algorithm~\cite{sinkhorn1967diagonal}, viewing SAA through the lens of optimal transport~\cite{cuturi2013sinkhorn}.

Given an initial blank ``canvas'' having $N = HW$ pixels, we define a uniform source histogram $\mv{r}={N^{-1}\boldsymbol{1}_N}$, standing for the equal chance of each pixel to be ``drawn'' or occupied by one of the classes.
The target histogram, or semantic palette, $\mv{t}\in\mathbb{R}^C_{+}$, defines the prescribed ``budget'' for the $C$ classes.
One can defined the set of admissible \textit{transport plans} from one distribution to the other one: 
\begin{align}
U(\mv{r},\mv{t}) := \{\mv{P} \in \mathbb{R}^{C\times N}_{+} | \mv{P}\boldsymbol{1}_N = \mv{t}, \mv{P}^{\top}\boldsymbol{1}_C & = \mv{r}\}.
\end{align}
A connection of the soft mask $\mv{m}$ with these transport plans is established as follows. Flattening spatial dimensions ($H\times W \rightarrow N$), the soft mask $\mv{m}$ is now in $[0,1]^{C\times N}$, and it is expected to simultaneously verify:
\begin{align}
    N^{-1}\mv{m}\boldsymbol{1}_N & = \mv{t}\, ,\label{eq:sinkhorn11}\\
    N^{-1}\mv{m}^{\top}\boldsymbol{1}_C & = \mv{r}\, ,\label{eq:sinkhorn22}
\end{align}
where (\ref{eq:sinkhorn11}) warrants that soft pixel-to-class assignments respect the input class proportions ($\frac{1}{N}\sum_n \mv{m}_{c,n} = \mv{t}_c$) and (\ref{eq:sinkhorn22}) ensures that at each pixel location there is a valid class distribution ($\sum_c \mv{m}_{c,n} = 1$). If $\mv{m}$ verifies both, then $N^{-1}\mv{m}\in U(\mv{r},\mv{t})$. Note that, in practice, only (\ref{eq:sinkhorn22}) is a hard constraint. 

We can now formulate the task of finding $\frac{1}{N}\mv{m}$ as solving an entropy-regularized optimal-transport problem \cite{cuturi2013sinkhorn}:
\begin{align}
\mv{P}^{*} = \underset{\mv{P} \in U(\mv{r},\mv{t})}{\mathrm{argmin}} \langle \mv{P},\mv{K}\rangle - \frac{1}{\lambda}h(\mv{P}),
\end{align}
where $\mv{K}\in \mathbb{R}^{C\times N}$ is a suitable transport-cost matrix, $h(\mv{P})$ is the entropy of $\mv{P}$, $\lambda$ is a weight (fixed as $1$ next) and $\langle\,,\rangle$ denotes the Frobenius dot-product. 

In the SAA module, the cost matrix $\mv{K}$ is defined as $-\mv{f}$. Intuitively $\mv{f}$, the ``raw'' output of our network, indicates the initial class preference of each pixel $i$; its opposite $-\mv{f}$ can be seen as the transportation cost, \ie, the higher the chance to assign pixel $i$ to class $c$, the lower the cost to ``transport'' from pixel $i$ to class $c$ is.

To find the optimal plan $\mv{P}^{*}$, one can adopt the Sinkhorn algorithm, initializing $\mv{P}$ as $\exp(-\mv{K})=\exp(\mv{f})$ and alternating row-wise and column-wise normalization/scaling steps \cite{cuturi2013sinkhorn}:
\begin{align}
    \mv{P} ~\leftarrow ~ & \mathrm{diag}\big[\mv{t} \oslash (\mv{P}\boldsymbol{1}_N)\big] \mv{P}\, , \label{eq:row1}\\ 
    \mv{P} ~\leftarrow ~ & \mv{P} \mathrm{diag}\big[\mv{r} \oslash (\mv{P}^\top\boldsymbol{1}_C)\big]\, , \label{eq:col1}
\end{align}
where $\oslash$ denotes the Hadamard entry-wise division. Eq.\,\ref{eq:row1} amounts to successively normalizing each of the $C$ rows and then multiplying each by its target probability in $\mv{t}$ -- exactly how $\wdis$ is derived from $\mv f$;
Eq.\,\ref{eq:col1} amounts to normalizing each of the $N$ columns -- exactly how $\mask$ is derived from $\wdis$ (since $\mask$ corresponds to $N\mv{P}$).

Effectively, the steps of the SAA presented in Section~\ref{sec:layout_arch} correspond to a single step of this Sinkhorn algorithm. Having more steps is possible, yet we opted to a single one as to allow certain slacks in the final scene composition, \ie, not forcing an exact matching to the input semantic palette.

\section{Direct matching loss in Baseline 1}
\label{app:direct_matching_loss}
The \textit{baseline 1} introduced in Section~\ref{sec:exp-cond} uses a direct matching loss to enforce conditioning constraints.
We provide here the detail of this loss.

The conditional layout generator $G$
produces semantic soft probability masks $\mask\in [0,1]^{C\times H\times W}$.
Let us define $\phi: [0,1]^{C\times H\times W}\rightarrow\Delta^{C}$ the function that computes the class histogram of the final semantic map derived from soft mask $\mv{m}$, where
$\Delta^{C} := \{\mv{x}\in\mathbb{R}^C_{+}: \mv{x}^{\top}\boldsymbol{1}_C=1\}$ is the probability simplex. 
For each class $c\in\llbracket 1, C \rrbracket$,
the proportion of pixels
assigned to this class in the image is given by:
\begin{equation}
	\phi_c(\mask) = \frac{1}{HW} \sum_{(i,j) \in \Omega} [\,{\underset{k}{\mathrm{argmax}} \, \mask_{k,i,j}} = c\,].
	\label{eq:proportions}
\end{equation}
This function $\phi$ being non-differentiable, it cannot be easily used to define a training loss. 
Instead, we propose to use 
$\widehat{\phi}$, a differentiable \textit{soft} estimation of the semantic histogram, defined as:
\begin{equation}
	\widehat{\phi}_c(\mask) = \frac{1}{HW} \sum_{(i,j) \in \Omega} \mask_{c,i,j},
\end{equation}
for each $c\in\llbracket 1, C \rrbracket$. The matching loss in \textit{baseline 1} is finally defined as the KL-divergence between target and estimated histograms:
\begin{equation}
	\mathcal{L}_{\mathrm{MATCH}}(G) = \mathbb{E}_{(\mv{z},\mv{t})} 
	\Big[\sum_{c \in \llbracket 1, C \rrbracket} 
	\!\mv{t}_c \cdot
	\log\Big(\frac{\mv{t}_c}{\widehat{\phi}_c(G(\mv{z},\mv{t}))}\Big)\Big].
	\label{eq:sem_recover}
\end{equation}

\section{Domain adaptation for data augmentation}
\label{app:da_for_augmentation}
We explain here how AdvEnt, the domain-adaptation technique in~\cite{vu2019advent}, is mobilized in Section~\ref{sec:exp_data_augmentation} when using both real and synthetic data. 
In effect, we adopt the main ingredient of AdvEnt: an adversarial training procedure to perform alignment on the so-called weighted self-information space. While the segmenter is trained as usual, an additional discriminator, taking segmenter's prediction as input, is trained in parallel to determine from which domain (real or synthetic) the prediction originates.
Playing the adversarial game, the segmenter tries to fool the discriminator, eventually resulting in closing the domain gap.

Such a technique has been proven effective for unsupervised domain adaptation in semantic segmentation, where images are annotated only in one domain.
We revisit it in a different context, where full annotations are available in both domains.
Empirical results in Table~\ref{tab:data_augmentation} demonstrate the benefit of addressing domain gap this way  when using synthesized data for data augmentation.
Since for DA we use the default hyper-parameters from ~\cite{vu2019advent}, 
fine-tuning them might yield even higher performance.

\section{Novelty loss for face editing}
\label{app:novelty_loss}
In the case of partial editing, the conditional layout generator $G$ takes a semantic layout $\mv l$ as input, in addition to the noise $\mv z$ and the target palette $\mv t$. We denote  $G(\mv{z},\mv{t}, \mv{l})=\mv{m}$ the final edited layout produced by the generator, after the generated partial layout and the input layout have been merged. For the face editing task, different from the partial-editing method proposed for data augmentation in urban scenes, the input layout is not cropped. In addition,
we introduce a novelty loss on top of the conditional and adversarial losses, to ensure that the edits do modify the original content. It is defined by:
\begin{equation}
	\mathcal{L}_{\mathrm{NOV}}(G) = \frac{1}{|E|} \mathbb{E}_{(\mv{z},\mv{t},\mv{l})} \Big[\sum_{(i,j) \in E} \sum_{c \in \llbracket 1, C \rrbracket} \mv{l}_{c,i,j}\mv{m}_{c,i,j} \Big],
	\label{eq:sem_novelty}
\end{equation}
where $E\subset\Omega$ is the set of pixel locations where an edit has been made, \ie, the dominant class in the partial layout is not the \textit{background} class. This loss is, at every edited pixel location, the scalar product between the generated soft probability distribution and the input one-hot one, and therefore promotes orthogonal content between the two.

\section{Better base generative frameworks}
\label{app:discussion_better_gan}
In this work, we built the layout synthesizer
upon ProGAN~\cite{karras2017progressive} as to guarantee a fair comparison to SBGAN~\cite{azadi2019semantic} and to highlight the merits of the proposed architecture designs and learning objectives.
We note that this part of the Semantic Palette’s pipeline can leverage any other
hierarchical GAN architecture, for example StyleGAN~\cite{karras2019style} or StyleGAN2~\cite{karras2020analyzing}.
In fact, the choice of the base generative framework is orthogonal to our contributions and any improvement on it should increase the performance of both
the Semantic Palette and the considered baselines.
Similarly, for the image generation part, we adopted GauGAN~\cite{park2019semantic} as done
in SB-GAN~\cite{azadi2019semantic} to ensure fair comparison while noting that the
choice of this generator is orthogonal to our contributions; In particular, if using a different framework like CC-FPSE~\cite{liu2019learning} was to bring improvements, they would benefit all compared pipelines.

\section{Implementation details}
\label{app:impl_details}
\textbf{Weights for losses.} All the introduced losses are equally weighted in the experiments. However, a particular weighting may prove useful for specific applications such as to improve further the semantic control at the expense of a slight degradation of the image realism or the other way around.

\textbf{Layout synthesis model.} The layout generator is trained with ADAM~\cite{kingma2014adam}, an initial learning rate of $10^{-3}$, $\beta=(0,0.99)$ and specific epochs (600 to 150) and batch sizes (1024 to 8) for every resolution ($4\stimes8$ to $128\stimes256$). 

\textbf{Image synthesis model.} The image generator is trained with ADAM~\cite{kingma2014adam}, an initial learning rate of $2\cdot 10^{-4}$, $\beta=(0.5,0.999)$ for $200$ epochs and a batch size of $8$. 

\begin{table*}[ht!]
\setlength\heavyrulewidth{0.25ex}
\setlength{\tabcolsep}{0pt}
\aboverulesep=0ex
\belowrulesep=0ex
\small

\centering
\begin{tabular}{@{}l@{}|@{}l@{}|@{}c@{}|@{}L{1.5cm}@{}L{1.5cm}@{}|@{}L{1.5cm}@{}L{1.5cm}@{}|@{}L{1.5cm}@{}L{1.5cm}@{}}
\multicolumn{1}{c}{}&\multicolumn{1}{c}{}&\multicolumn{1}{c}{}&\multicolumn{2}{c}{\small \rule{0pt}{2.5ex}\textbf{(a) Cityscapes}}&\multicolumn{2}{c}{\small \rule{0pt}{2.5ex}\textbf{(b) Cityscapes-25k}}&\multicolumn{2}{c}{\small \rule{0pt}{2.5ex}\textbf{(c) IDD}}\\
\toprule
Data & \,\, Method & \,\,Crop\,\, & \,\, $\text{mIoU}^{*}$ & mIoU & \,\, $\text{mIoU}^{*}$ & mIoU & \,\, $\text{mIoU}^{*}$ & mIoU \\
\midrule
\multirow{2}{*}{Real} & \multirow{2}{*}{\,\, Baseline} &  & \,\, 36.9 & 48.1 & \,\, 36.5 & 53.0 & \,\,  33.8 & 43.8 \\
& &\cellcolor[gray]{.92} \checkmark &\cellcolor[gray]{.92} \,\, $35.7_{\,\down{1.2}}$ &\cellcolor[gray]{.92} $51.4_{\,\up{3.3}}$ &\cellcolor[gray]{.92} \,\, $35.5_{\,\down{1.0}}$ &\cellcolor[gray]{.92} $59.6_{\,\up{6.6}}$ &\cellcolor[gray]{.92} \,\,  $32.7_{\,\down{1.1}}$ &\cellcolor[gray]{.92} $40.0_{\,\down{3.8}}$ \\
\midrule
\multirow{4}{*}{Real + Syn\,\,} & \multirow{2}{*}{\,\, Sem. Palette (DA)} & & \,\, $38.6$ &$51.6$ &\,\, $38.6$ &$57.3$ &\,\, $34.5$ &$44.7$ \\
 & &\cellcolor[gray]{.92} \checkmark &\cellcolor[gray]{.92} \,\, $38.7_{\,\up{0.1}}$ &\cellcolor[gray]{.92}$52.2_{\,\up{0.6}}$ &\cellcolor[gray]{.92}\,\, $32.9_{\,\down{5.7}}$ &\cellcolor[gray]{.92}$57.0_{\,\down{0.3}}$ &\cellcolor[gray]{.92}\,\, $29.8_{\,\down{4.7}}$ &\cellcolor[gray]{.92}$38.5_{\,\down{6.2}}$ \\
\cmidrule(lr){2-9}
 &\multirow{2}{*}{\,\, Sem. Palette (Part. + DA) \,\, }& &\,\, $\textbf{40.7}$ &$52.6$ &\,\, $\textbf{42.5}$ &$\textbf{60.5}$ &\,\, $\textbf{35.3}$ &$\textbf{45.8}$ \\
 &&\cellcolor[gray]{.92} \checkmark &\cellcolor[gray]{.92}\,\, $39.8_{\,\down{0.9}}$ &\cellcolor[gray]{.92}$\textbf{54.4}_{\,\up{1.8}}$ &\cellcolor[gray]{.92}\,\, ${37.0}_{\,\down{5.5}}$ &\cellcolor[gray]{.92}$56.4_{\,\down{4.1}}$ &\cellcolor[gray]{.92}\,\, $31.1_{\,\down{4.2}}$ &\cellcolor[gray]{.92}$39.7_{\,\down{6.1}}$ \\

\bottomrule

\end{tabular}
\vspace{0.1cm}

\vspace{-0.2cm}
\caption{\small\textbf{Using cropping to augment real data.} Same notations as in Table~\ref{tab:data_augmentation}. In each group, models using cropping are compared against the ones without.}
\label{tab:data_augmentation_crop}
\vspace{-0.3cm}
\end{table*}

\textbf{Segmenter.} 
We train a DeeplabV3~\cite{chen2017rethinking} model with Stochastic Gradient Descent, an initial learning rate of $10^{-2}$, $0.9$ momentum, $5\cdot 10^{-4}$ weight decay, for $300$ epochs and a batch size of $16$.

\textbf{Palette generator.} The GMM model is trained on the semantic proportions from the real training dataset, using the expectation-maximization (EM) algorithm. The number of Gaussian components control the trade-off between approximation and generalization. We select their number using the Akaike Information Criterion (AIC), which balances these two objectives.

In practice, to ensure that the vectors sampled from the GMM are true proportions, \ie, non-negative and $L_1$-normalized, one has to project them onto the probability $C$--simplex.
As the projection is not easy to compute analytically, one can get a good approximation with constrained minimization methods such as the trust-region constrained algorithm.
However, their convergence is slow, making them impractical in our case.
Instead, we chose to compute a rough estimate of the projection by first clipping the sampled vectors to $[0, 1]$ and then normalizing them.

\section{Standard data augmentation.}
\label{app:comparison_standard_aug}
We used random horizontal flipping in all experiments done in the main paper.

Cropping is another standard data augmentation strategy used in semantic segmentation.
We provide here an ablation study where we additionally perform cropping to augment real data while training the baseline and Semantic Palette models. 
We report in Table~\ref{tab:data_augmentation_crop} the performance on the three benchmarks.
In terms of mIoU, we observe only on Cityscapes that cropping helps improve all methods and achieves best scores when combined with our augmentation strategy; yet, on the other datasets, having cropping degrades the performance.
In terms of mIoU$^*$, the performance drops in most cases.
These results reveal different behaviors in the three datasets when including cropping in the data augmentation procedure during training.
We note that, as cropping is done only on real data, increase or decrease in performance by using it is orthogonal to our proposed framework.
Overall, the best results are obtained using Semantic Palette.

\section{Additional experiments.}
\label{app:more_exp}
We provide results of a few additional experiments aimed at evaluating the Semantic Palette in different setups, namely with other types of data and at higher resolution.

\textbf{Effect of Semantic Palette on non-urban scenes.} To verify that the proposed method generalizes well to other types of natural images, we trained the Semantic Palette and the unconditional baselines on the ADE-Indoor dataset~\cite{zhou2017ade} at $128\stimes256$ resolution. The results, as shown in Table~\ref{tab:ade-indoor}, confirm the advantage of our model over the unconditional baselines.

\begin{table}
\setlength\heavyrulewidth{0.25ex}
\setlength{\tabcolsep}{0pt}
\aboverulesep=0ex
\belowrulesep=0ex
\small
    \centering
    \begin{tabular}{@{}l@{}|@{}C{1.45cm}@{}C{1.45cm}@{}C{1.45cm}@{}C{1.45cm}@{}}
    \toprule
    
    \multirow{2}{*}{Method} & Layout & Image & GAN-test & GAN-train \\
    \cmidrule(lr){2-2} \cmidrule(lr){3-3} \cmidrule(lr){4-4} \cmidrule(lr){5-5}
     & FSD $\downarrow$ & FID $\downarrow$ & mIoU &  mIoU\\
    \midrule
    PCGAN~\cite{howe2019conditional} & 211.7 & 96.6 & 11.1 & 6.8\\
    SB-GAN~\cite{azadi2019semantic} & 211.7 & 93.7 & 12.2 & 7.0\\
    Sem. Palette & \textbf{76.1} & \textbf{88.4} & \textbf{20.1} & \textbf{10.5}\\
    \midrule
    SB-GAN\textsubscript{\,e2e}~\cite{azadi2019semantic}\,\, & 93.3 & 82.7 & 15.3 & 10.0\\
    Sem. Palette\textsubscript{\,e2e}\,\, & \textbf{19.0} & \textbf{76.5} & \textbf{21.4} & \textbf{11.7}\\
    \bottomrule
    \end{tabular}
	\vspace{-0.3cm}
    \caption{\small\textbf{Results on ADE-Indoor}.}
    \label{tab:ade-indoor}
    \vspace{-0.3cm}
\end{table}

\textbf{Ability to scale to higher resolutions.} To afford an extensive evaluation of the proposed methods compared to the different baselines, all experiments were conducted at the $128\stimes256$ resolution. To evaluate the performance at higher resolution, we now compare the Semantic Palette to SB-GAN at the $256\stimes512$ resolution on Cityscapes. The results, in Table~\ref{tab:higher-res}, turn out to be consistent with the ones reported at $128\stimes256$.

\begin{table}
\setlength\heavyrulewidth{0.25ex}
\setlength{\tabcolsep}{0pt}
\aboverulesep=0ex
\belowrulesep=0ex
\small
    \centering
    \begin{tabular}{@{}l@{}|@{}C{1.05cm}@{}C{1.05cm}@{}C{1.05cm}@{}C{1.05cm}@{}C{1.05cm}@{}C{1.05cm}@{}}
    \toprule
    
    \multirow{2}{*}{Method} & Layout & Image & \multicolumn{2}{c}{GAN-test} & \multicolumn{2}{c}{GAN-train} \\
    \cmidrule(lr){2-2} \cmidrule(lr){3-3} \cmidrule(lr){4-5} \cmidrule(lr){6-7}
     & FSD $\downarrow$ & FID $\downarrow$ & $\text{mIoU}^{*}$ &  mIoU & $\text{mIoU}^{*}$ &  mIoU\\
    \midrule
    SB-GAN~\cite{azadi2019semantic}\,\, & 66.0 & 74.8 & 34.9 & 46.0 & 28.0 & 35.7\\
    Sem. Palette\,\, & \textbf{32.4} & \textbf{66.7} & \textbf{38.0} & \textbf{50.0} & \textbf{32.1} & \textbf{42.3}\\
    \bottomrule
    \end{tabular}
	\vspace{-0.3cm}
    \caption{\small\textbf{Results on Cityscapes at resolution $256\stimes512$}.}
    \label{tab:higher-res}
    \vspace{-0.3cm}
\end{table}

\section{Qualitative results}
\label{app:more_quals}
\begin{figure*}
	\setlength\tabcolsep{1.0pt}
	\centering
	\begin{tabular}{lccccc}
	    \begin{adjustbox}{valign=t, raise=-1ex}
			\includegraphics[height=.15\linewidth]{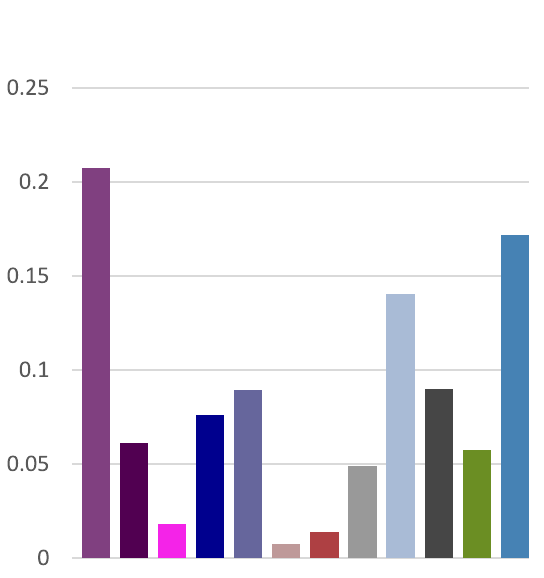}
		\end{adjustbox}
		&
		\begin{adjustbox}{valign=t}
			\begin{tabular}{@{}c@{}}
				\includegraphics[width=.15\linewidth]{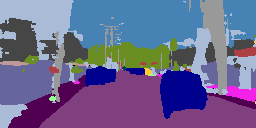}\\[-0.5mm]
				\includegraphics[width=.15\linewidth]{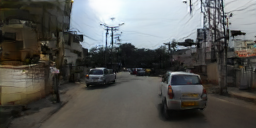}
			\end{tabular}
		\end{adjustbox}
		&
		\begin{adjustbox}{valign=t}
			\begin{tabular}{@{}c@{}}
				\includegraphics[width=.15\linewidth]{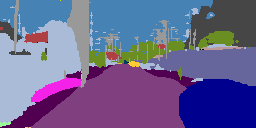}\\[-0.5mm]
				\includegraphics[width=.15\linewidth]{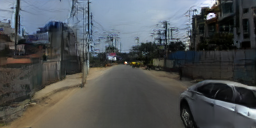}
			\end{tabular}
		\end{adjustbox}
		&
		\begin{adjustbox}{valign=t}
			\begin{tabular}{@{}c@{}}
				\includegraphics[width=.15\linewidth]{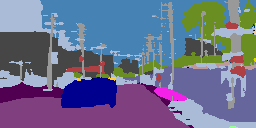}\\[-0.5mm]
				\includegraphics[width=.15\linewidth]{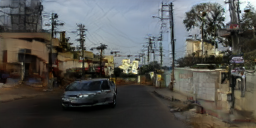}
			\end{tabular}
		\end{adjustbox}
		&
		\begin{adjustbox}{valign=t}
			\begin{tabular}{@{}c@{}}
				\includegraphics[width=.15\linewidth]{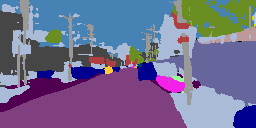}\\[-0.5mm]
				\includegraphics[width=.15\linewidth]{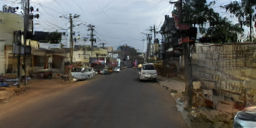}
			\end{tabular}
		\end{adjustbox}
		&
		\begin{adjustbox}{valign=t}
			\begin{tabular}{@{}c@{}}
				\includegraphics[width=.15\linewidth]{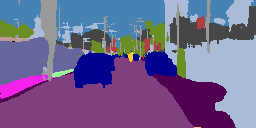}\\[-0.5mm]
				\includegraphics[width=.15\linewidth]{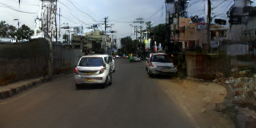}
			\end{tabular}
		\end{adjustbox}\vspace{-0.25em}\\ 
		
		\midrule
		
		\begin{adjustbox}{valign=t, raise=-1ex}
			\includegraphics[height=.15\linewidth]{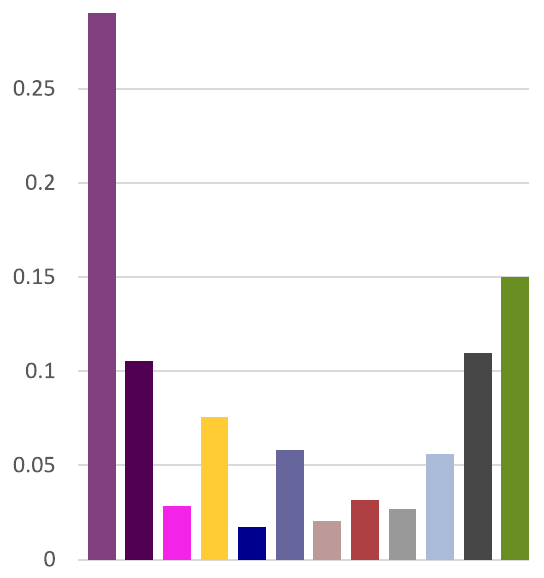}
		\end{adjustbox}
		&
		\begin{adjustbox}{valign=t}
			\begin{tabular}{@{}c@{}}
				\includegraphics[width=.15\linewidth]{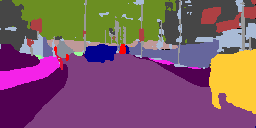}\\[-0.5mm]
				\includegraphics[width=.15\linewidth]{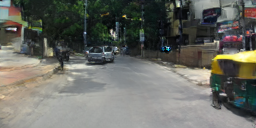}
			\end{tabular}
		\end{adjustbox}
		&
		\begin{adjustbox}{valign=t}
			\begin{tabular}{@{}c@{}}
				\includegraphics[width=.15\linewidth]{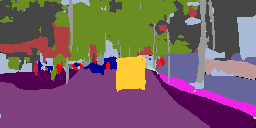}\\[-0.5mm]
				\includegraphics[width=.15\linewidth]{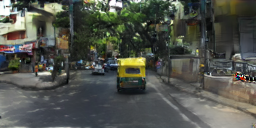}
			\end{tabular}
		\end{adjustbox}
		&
		\begin{adjustbox}{valign=t}
			\begin{tabular}{@{}c@{}}
				\includegraphics[width=.15\linewidth]{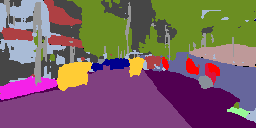}\\[-0.5mm]
				\includegraphics[width=.15\linewidth]{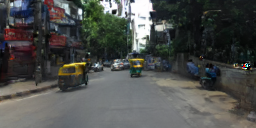}
			\end{tabular}
		\end{adjustbox}
		&
		\begin{adjustbox}{valign=t}
			\begin{tabular}{@{}c@{}}
				\includegraphics[width=.15\linewidth]{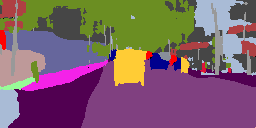}\\[-0.5mm]
				\includegraphics[width=.15\linewidth]{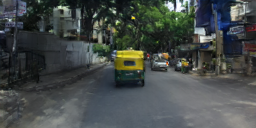}
			\end{tabular}
		\end{adjustbox}
		&
		\begin{adjustbox}{valign=t}
			\begin{tabular}{@{}c@{}}
				\includegraphics[width=.15\linewidth]{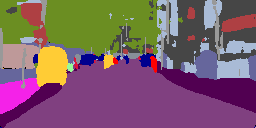}\\[-0.5mm]
				\includegraphics[width=.15\linewidth]{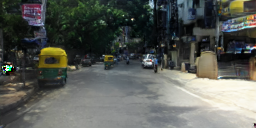}
			\end{tabular}
		\end{adjustbox}\vspace{-0.25em}\\ 
		
		\midrule
		
		\begin{adjustbox}{valign=t}
			\includegraphics[height=.16\linewidth]{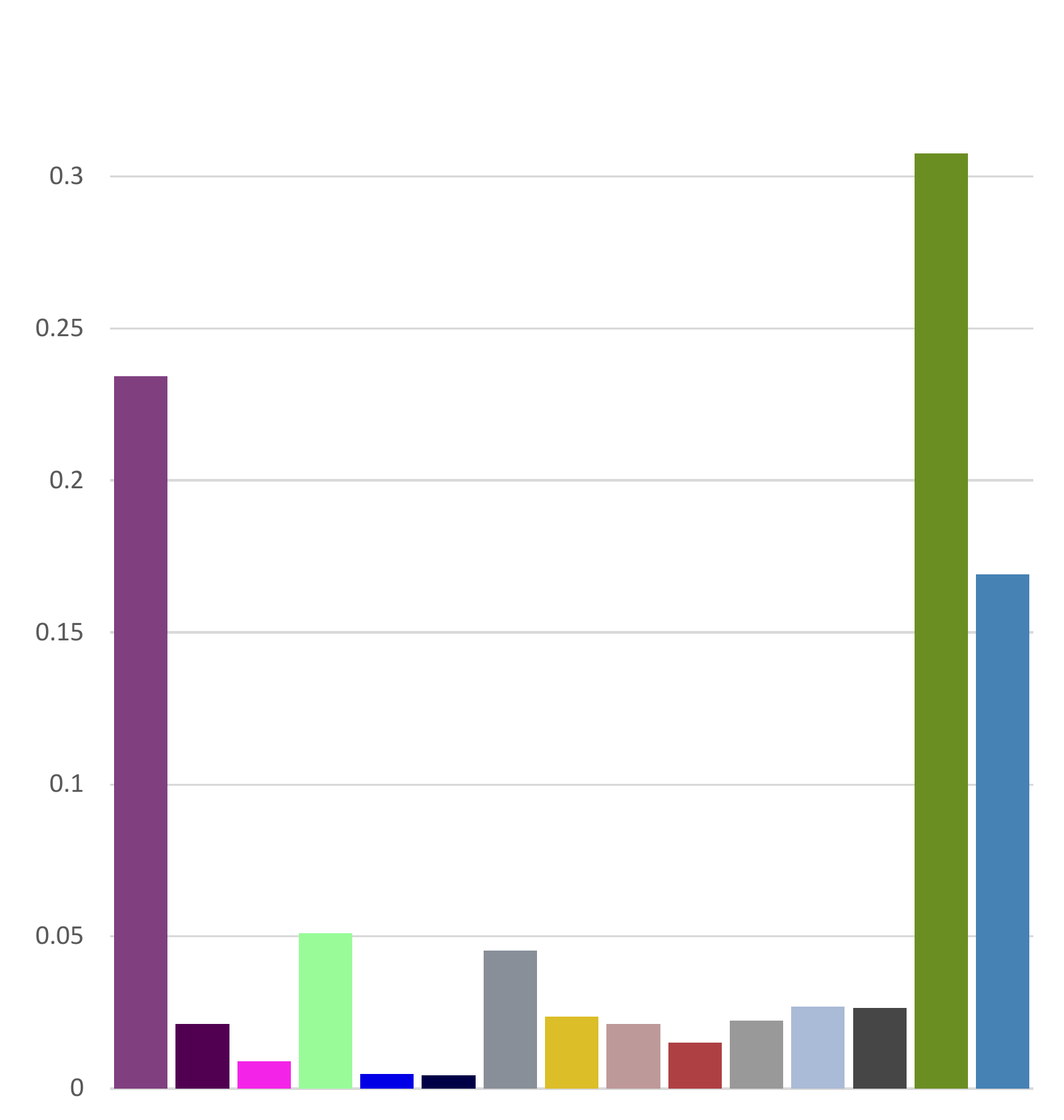}
		\end{adjustbox}
		&
		\begin{adjustbox}{valign=t}
			\begin{tabular}{@{}c@{}}
				\includegraphics[width=.15\linewidth]{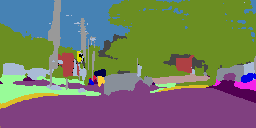}\\[-0.5mm]
				\includegraphics[width=.15\linewidth]{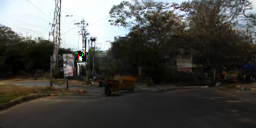}
			\end{tabular}
		\end{adjustbox}
		&
		\begin{adjustbox}{valign=t}
			\begin{tabular}{@{}c@{}}
				\includegraphics[width=.15\linewidth]{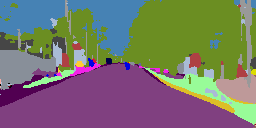}\\[-0.5mm]
				\includegraphics[width=.15\linewidth]{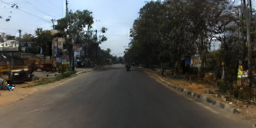}
			\end{tabular}
		\end{adjustbox}
		&
		\begin{adjustbox}{valign=t}
			\begin{tabular}{@{}c@{}}
				\includegraphics[width=.15\linewidth]{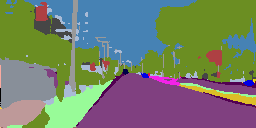}\\[-0.5mm]
				\includegraphics[width=.15\linewidth]{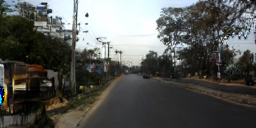}
			\end{tabular}
		\end{adjustbox}
		&
		\begin{adjustbox}{valign=t}
			\begin{tabular}{@{}c@{}}
				\includegraphics[width=.15\linewidth]{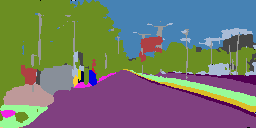}\\[-0.5mm]
				\includegraphics[width=.15\linewidth]{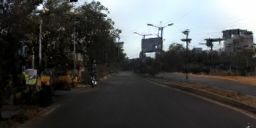}
			\end{tabular}
		\end{adjustbox}
		&
		\begin{adjustbox}{valign=t}
			\begin{tabular}{@{}c@{}}
				\includegraphics[width=.15\linewidth]{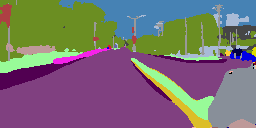}\\[-0.5mm]
				\includegraphics[width=.15\linewidth]{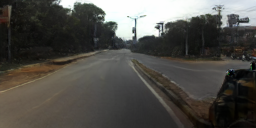}
			\end{tabular}
		\end{adjustbox}\vspace{-0.25em}\\ 
		
		\midrule
		
		\begin{adjustbox}{valign=t}
			\includegraphics[height=.16\linewidth]{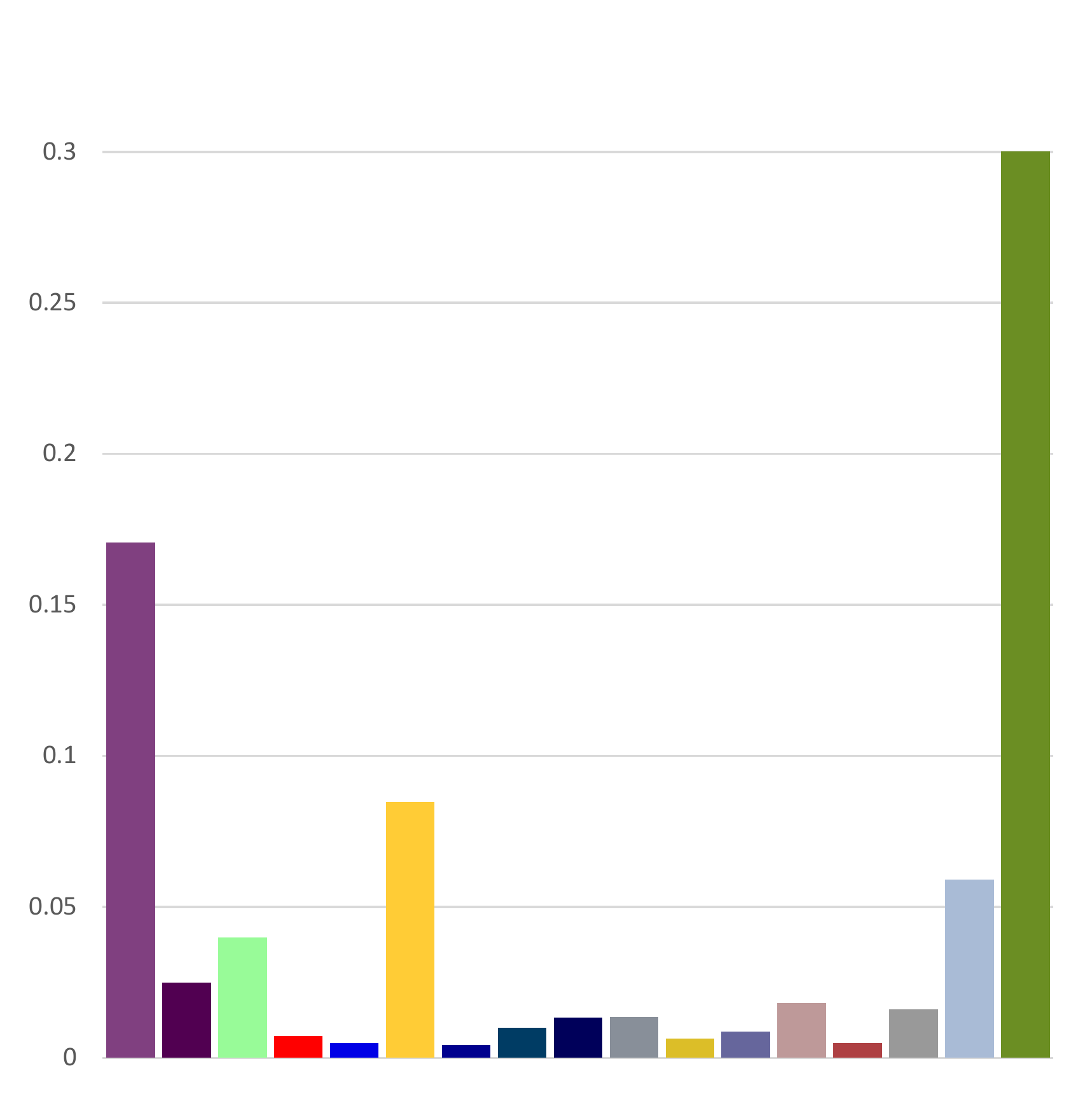}
		\end{adjustbox}
		&
		\begin{adjustbox}{valign=t}
			\begin{tabular}{@{}c@{}}
				\includegraphics[width=.15\linewidth]{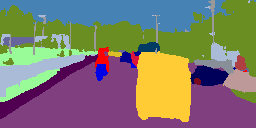}\\[-0.5mm]
				\includegraphics[width=.15\linewidth]{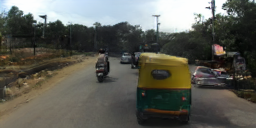}
			\end{tabular}
		\end{adjustbox}
		&
		\begin{adjustbox}{valign=t}
			\begin{tabular}{@{}c@{}}
				\includegraphics[width=.15\linewidth]{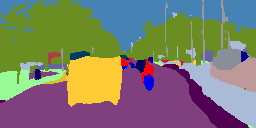}\\[-0.5mm]
				\includegraphics[width=.15\linewidth]{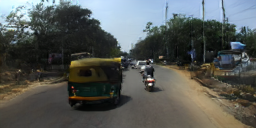}
			\end{tabular}
		\end{adjustbox}
		&
		\begin{adjustbox}{valign=t}
			\begin{tabular}{@{}c@{}}
				\includegraphics[width=.15\linewidth]{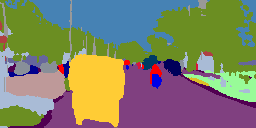}\\[-0.5mm]
				\includegraphics[width=.15\linewidth]{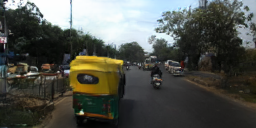}
			\end{tabular}
		\end{adjustbox}
		&
		\begin{adjustbox}{valign=t}
			\begin{tabular}{@{}c@{}}
				\includegraphics[width=.15\linewidth]{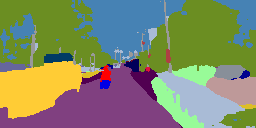}\\[-0.5mm]
				\includegraphics[width=.15\linewidth]{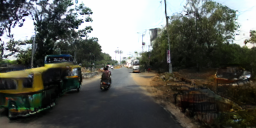}
			\end{tabular}
		\end{adjustbox}
		&
		\begin{adjustbox}{valign=t}
			\begin{tabular}{@{}c@{}}
				\includegraphics[width=.15\linewidth]{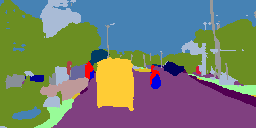}\\[-0.5mm]
				\includegraphics[width=.15\linewidth]{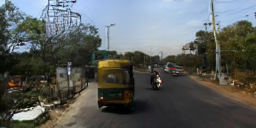}
			\end{tabular}
		\end{adjustbox}\vspace{-0.25em}\\ 
		
		\midrule
		
		\begin{adjustbox}{valign=t}
			\includegraphics[height=.16\linewidth]{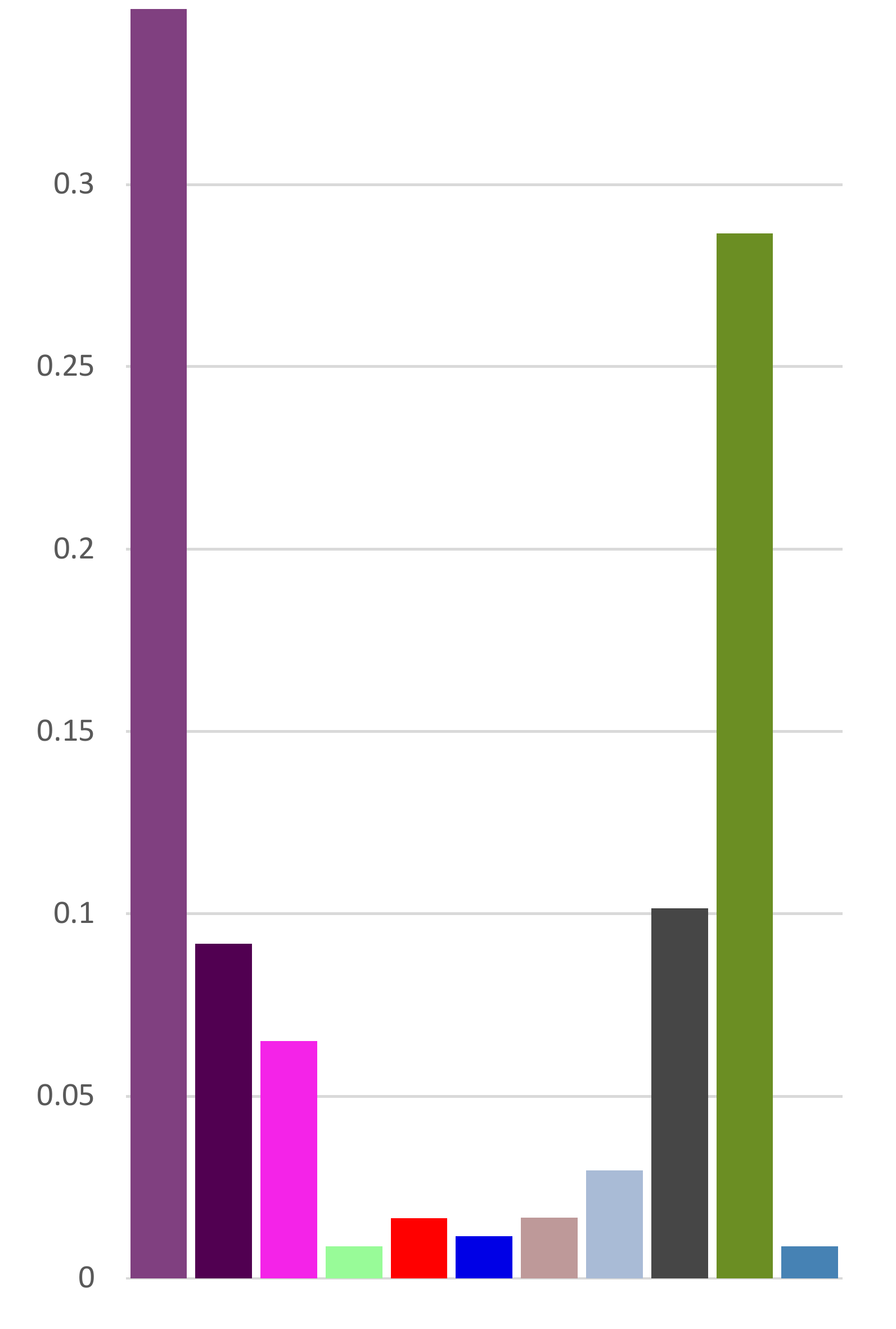}
		\end{adjustbox}
		&
		\begin{adjustbox}{valign=t}
			\begin{tabular}{@{}c@{}}
				\includegraphics[width=.15\linewidth]{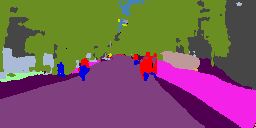}\\[-0.5mm]
				\includegraphics[width=.15\linewidth]{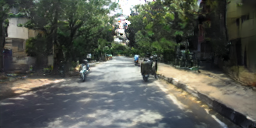}
			\end{tabular}
		\end{adjustbox}
		&
		\begin{adjustbox}{valign=t}
			\begin{tabular}{@{}c@{}}
				\includegraphics[width=.15\linewidth]{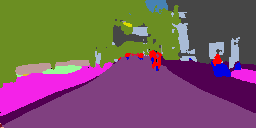}\\[-0.5mm]
				\includegraphics[width=.15\linewidth]{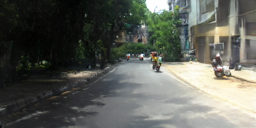}
			\end{tabular}
		\end{adjustbox}
		&
		\begin{adjustbox}{valign=t}
			\begin{tabular}{@{}c@{}}
				\includegraphics[width=.15\linewidth]{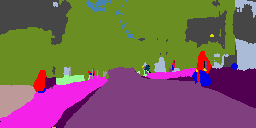}\\[-0.5mm]
				\includegraphics[width=.15\linewidth]{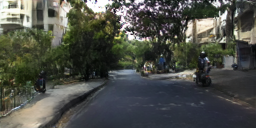}
			\end{tabular}
		\end{adjustbox}
		&
		\begin{adjustbox}{valign=t}
			\begin{tabular}{@{}c@{}}
				\includegraphics[width=.15\linewidth]{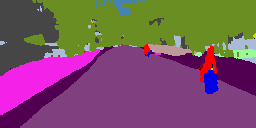}\\[-0.5mm]
				\includegraphics[width=.15\linewidth]{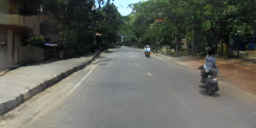}
			\end{tabular}
		\end{adjustbox}
		&
		\begin{adjustbox}{valign=t}
			\begin{tabular}{@{}c@{}}
				\includegraphics[width=.15\linewidth]{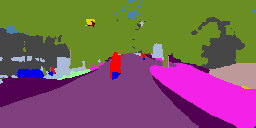}\\[-0.5mm]
				\includegraphics[width=.15\linewidth]{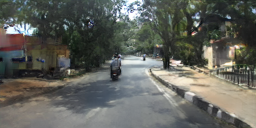}
			\end{tabular}
		\end{adjustbox}\vspace{-0.25em}\\ 
		
		\midrule
		
		\begin{adjustbox}{valign=t}
			\includegraphics[height=.16\linewidth]{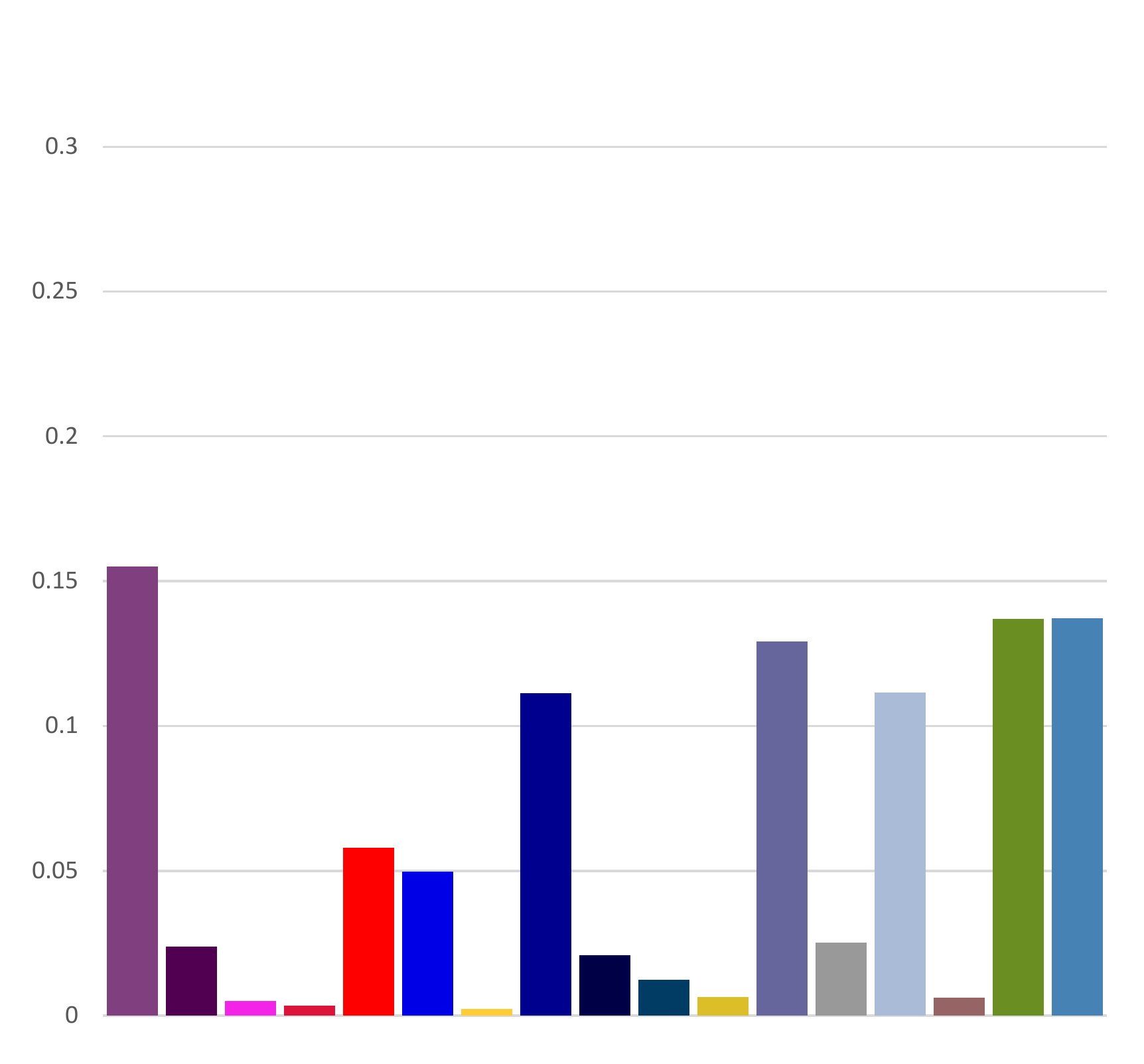}
		\end{adjustbox}
		&
		\begin{adjustbox}{valign=t}
			\begin{tabular}{@{}c@{}}
				\includegraphics[width=.15\linewidth]{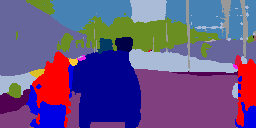}\\[-0.5mm]
				\includegraphics[width=.15\linewidth]{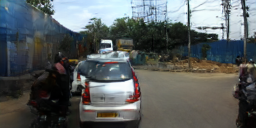}
			\end{tabular}
		\end{adjustbox}
		&
		\begin{adjustbox}{valign=t}
			\begin{tabular}{@{}c@{}}
				\includegraphics[width=.15\linewidth]{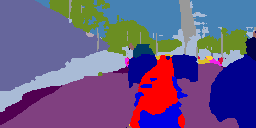}\\[-0.5mm]
				\includegraphics[width=.15\linewidth]{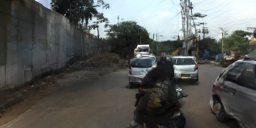}
			\end{tabular}
		\end{adjustbox}
		&
		\begin{adjustbox}{valign=t}
			\begin{tabular}{@{}c@{}}
				\includegraphics[width=.15\linewidth]{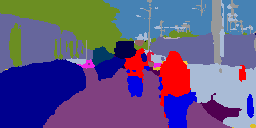}\\[-0.5mm]
				\includegraphics[width=.15\linewidth]{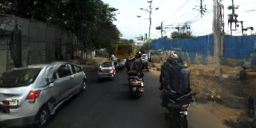}
			\end{tabular}
		\end{adjustbox}
		&
		\begin{adjustbox}{valign=t}
			\begin{tabular}{@{}c@{}}
				\includegraphics[width=.15\linewidth]{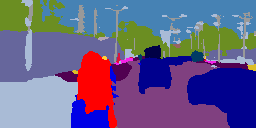}\\[-0.5mm]
				\includegraphics[width=.15\linewidth]{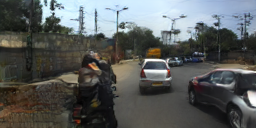}
			\end{tabular}
		\end{adjustbox}
		&
		\begin{adjustbox}{valign=t}
			\begin{tabular}{@{}c@{}}
				\includegraphics[width=.15\linewidth]{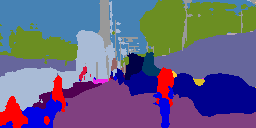}\\[-0.5mm]
				\includegraphics[width=.15\linewidth]{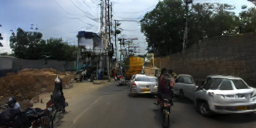}
			\end{tabular}
		\end{adjustbox}\vspace{-0.25em}\\ 

	\end{tabular}
	
	\vspace{-0.3cm}
	\caption{\small\textbf{Conditional layout-and-scene generation.} Various layout-scene pairs sampled from the same semantic code (left). 
	}
	\label{fig:idd}
	\vspace{-0.3cm}
\end{figure*}

\begin{figure*}
	\setlength\tabcolsep{1.0pt}
	\centering
	\begin{tabular}{cccc}
		(a) input layout & (b) cropped layout & (c) generated partial layout & (d) final merged layout \\
		
		\includegraphics[width=.24\linewidth]{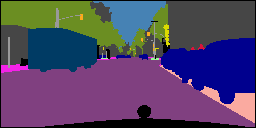} & 
		\includegraphics[width=.24\linewidth]{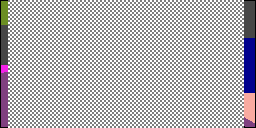} &
		\includegraphics[width=.24\linewidth]{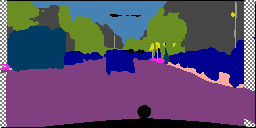} &
		\includegraphics[width=.24\linewidth]{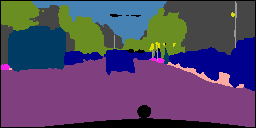} \\
		
		\includegraphics[width=.24\linewidth]{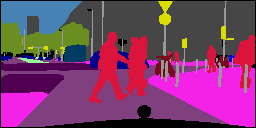} & 
		\includegraphics[width=.24\linewidth]{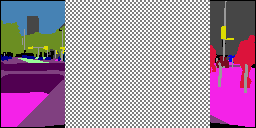} &
		\includegraphics[width=.24\linewidth]{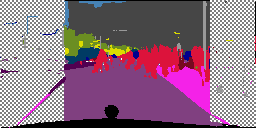} &
		\includegraphics[width=.24\linewidth]{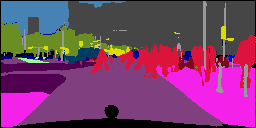} \\
		
		\includegraphics[width=.24\linewidth]{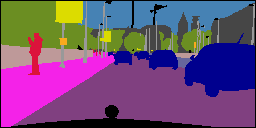} & 
		\includegraphics[width=.24\linewidth]{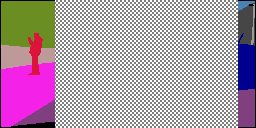} &
		\includegraphics[width=.24\linewidth]{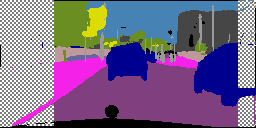} &
		\includegraphics[width=.24\linewidth]{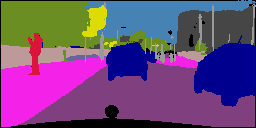} \\
		
		\includegraphics[width=.24\linewidth]{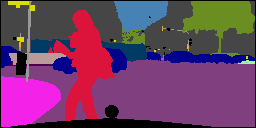} & 
		\includegraphics[width=.24\linewidth]{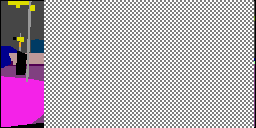} &
		\includegraphics[width=.24\linewidth]{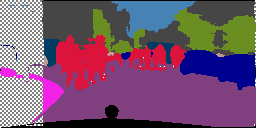} &
		\includegraphics[width=.24\linewidth]{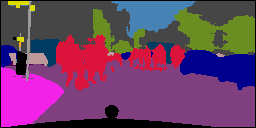} \\
		
		\includegraphics[width=.24\linewidth]{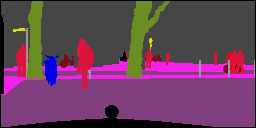} & 
		\includegraphics[width=.24\linewidth]{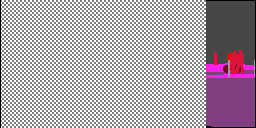} &
		\includegraphics[width=.24\linewidth]{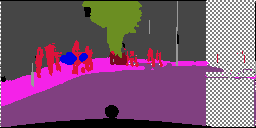} &
		\includegraphics[width=.24\linewidth]{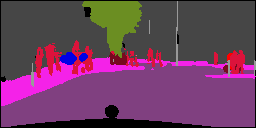} \\
		
		\includegraphics[width=.24\linewidth]{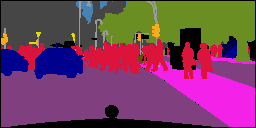} & 
		\includegraphics[width=.24\linewidth]{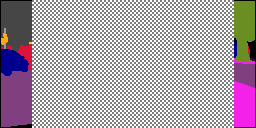} &
		\includegraphics[width=.24\linewidth]{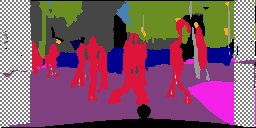} &
		\includegraphics[width=.24\linewidth]{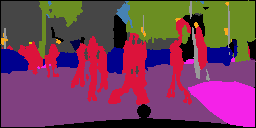} \\
		
		\includegraphics[width=.24\linewidth]{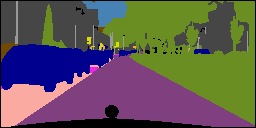} & 
		\includegraphics[width=.24\linewidth]{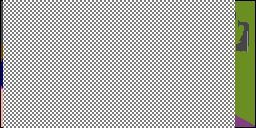} &
		\includegraphics[width=.24\linewidth]{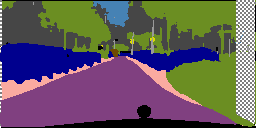} &
		\includegraphics[width=.24\linewidth]{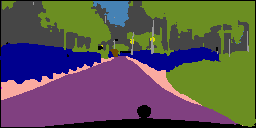} \\
		
		\includegraphics[width=.24\linewidth]{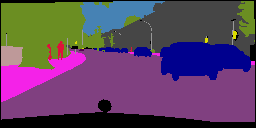} & 
		\includegraphics[width=.24\linewidth]{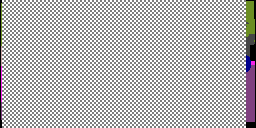} &
		\includegraphics[width=.24\linewidth]{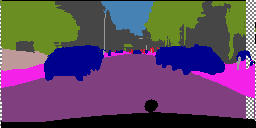} &
		\includegraphics[width=.24\linewidth]{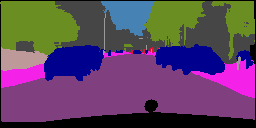} \\
		
		\includegraphics[width=.24\linewidth]{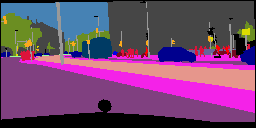} & 
		\includegraphics[width=.24\linewidth]{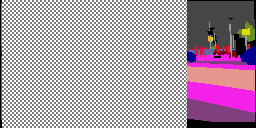} &
		\includegraphics[width=.24\linewidth]{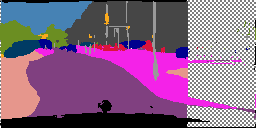} &
		\includegraphics[width=.24\linewidth]{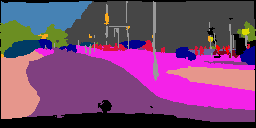} \\
	\end{tabular}
	\vspace{-0.3cm}
    \caption{\small\textbf{Partial editing of layouts.} The procedure consists in cropping ground-truth layouts and then synthesizing new objects within the cropped area, guided by the initial semantic proportions.}
	\label{fig:city}
	\vspace{-0.3cm}
\end{figure*}

\begin{figure*}
	\setlength\tabcolsep{1.0pt}
	\centering
	\begin{tabular}{cccccc}
		\includegraphics[width=.13\linewidth]{./figs/suppmat/hist/hair_1.pdf} &
		\includegraphics[width=.13\linewidth]{./figs/suppmat/hair0/part1.png} & 
		\includegraphics[width=.13\linewidth]{./figs/suppmat/hair0/part2.png} & 
		\includegraphics[width=.13\linewidth]{./figs/suppmat/hair0/part3.png} & 
		\includegraphics[width=.13\linewidth]{./figs/suppmat/hair0/part4.png} & 
		\includegraphics[width=.13\linewidth]{./figs/suppmat/hair0/part5.png} \\[-0.5mm]
		
		\includegraphics[width=.13\linewidth]{./figs/suppmat/hair0/rlay.png} &
		\includegraphics[width=.13\linewidth]{./figs/suppmat/hair0/glay1.png} & 
		\includegraphics[width=.13\linewidth]{./figs/suppmat/hair0/glay2.png} & 
		\includegraphics[width=.13\linewidth]{./figs/suppmat/hair0/glay3.png} & 
		\includegraphics[width=.13\linewidth]{./figs/suppmat/hair0/glay4.png} & 
		\includegraphics[width=.13\linewidth]{./figs/suppmat/hair0/glay5.png} \\[-0.5mm]
		
		\includegraphics[width=.13\linewidth]{./figs/suppmat/hair0/rimg.png} &
		\includegraphics[width=.13\linewidth]{./figs/suppmat/hair0/fimg1.png} &
		\includegraphics[width=.13\linewidth]{./figs/suppmat/hair0/fimg2.png} &
		\includegraphics[width=.13\linewidth]{./figs/suppmat/hair0/fimg3.png} &
		\includegraphics[width=.13\linewidth]{./figs/suppmat/hair0/fimg4.png} &
		\includegraphics[width=.13\linewidth]{./figs/suppmat/hair0/fimg5.png} \\ [-0.5mm]
	\end{tabular}
	
	\vspace{1mm}
	
	\begin{tabular}{cccccc}
		\includegraphics[width=.13\linewidth]{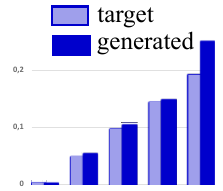} &
		\includegraphics[width=.13\linewidth]{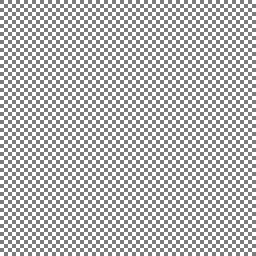} & 
		\includegraphics[width=.13\linewidth]{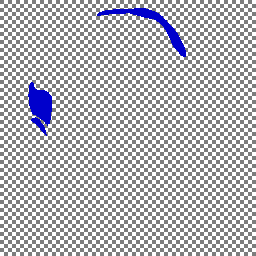} & 
		\includegraphics[width=.13\linewidth]{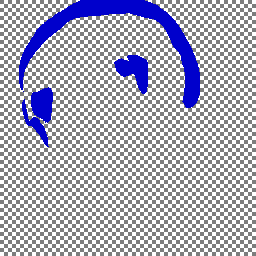} & 
		\includegraphics[width=.13\linewidth]{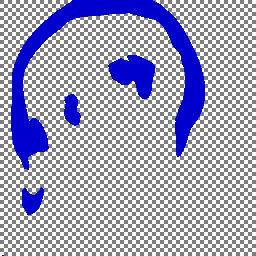} & 
		\includegraphics[width=.13\linewidth]{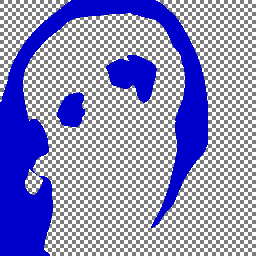} \\[-0.5mm]
		
		\includegraphics[width=.13\linewidth]{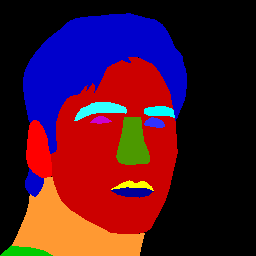} &
		\includegraphics[width=.13\linewidth]{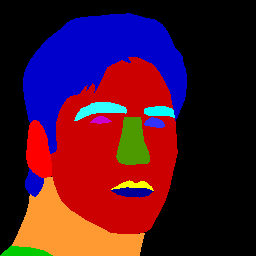} & 
		\includegraphics[width=.13\linewidth]{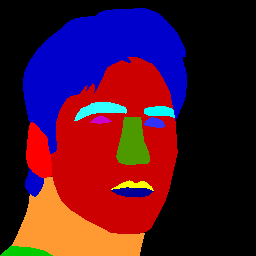} & 
		\includegraphics[width=.13\linewidth]{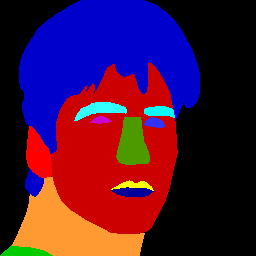} & 
		\includegraphics[width=.13\linewidth]{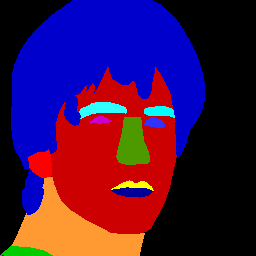} & 
		\includegraphics[width=.13\linewidth]{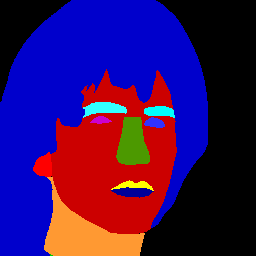} \\[-0.5mm]
		
		\includegraphics[width=.13\linewidth]{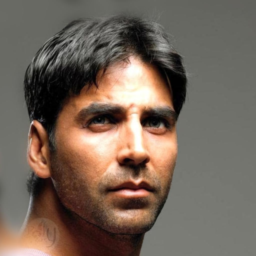} &
		\includegraphics[width=.13\linewidth]{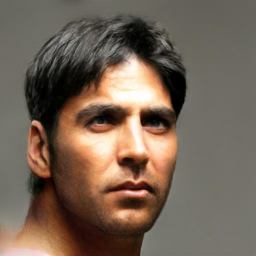} &
		\includegraphics[width=.13\linewidth]{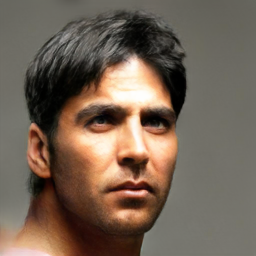} &
		\includegraphics[width=.13\linewidth]{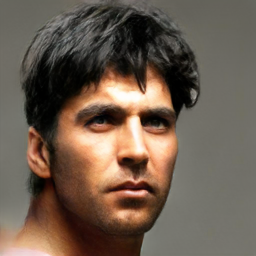} &
		\includegraphics[width=.13\linewidth]{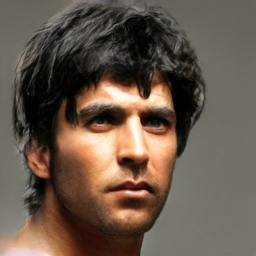} &
		\includegraphics[width=.13\linewidth]{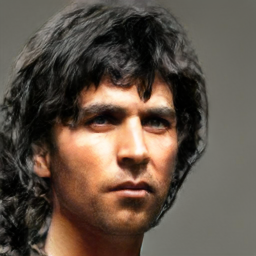} \\ [-0.5mm]
	\end{tabular}
	
	\vspace{1mm}
	
	\begin{tabular}{cccccc}
		\includegraphics[width=.13\linewidth]{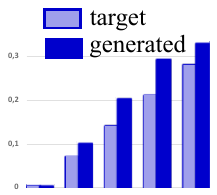} &
		\includegraphics[width=.13\linewidth]{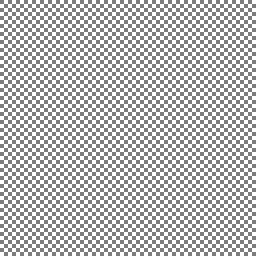} & 
		\includegraphics[width=.13\linewidth]{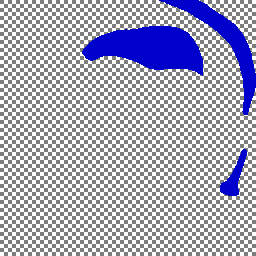} & 
		\includegraphics[width=.13\linewidth]{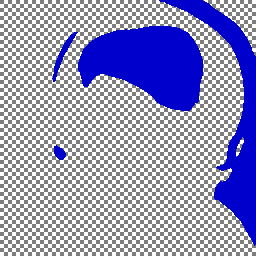} & 
		\includegraphics[width=.13\linewidth]{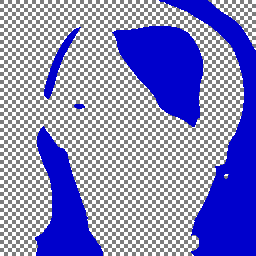} & 
		\includegraphics[width=.13\linewidth]{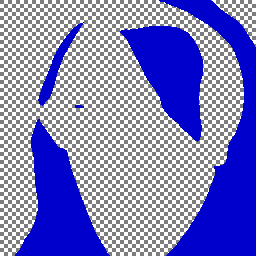} \\[-0.5mm]
		
		\includegraphics[width=.13\linewidth]{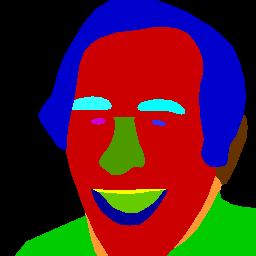} &
		\includegraphics[width=.13\linewidth]{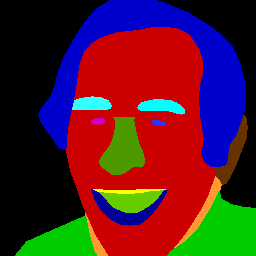} & 
		\includegraphics[width=.13\linewidth]{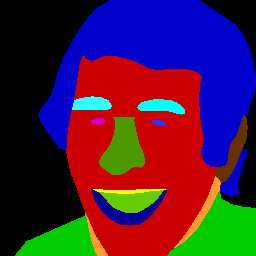} & 
		\includegraphics[width=.13\linewidth]{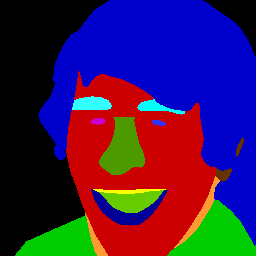} & 
		\includegraphics[width=.13\linewidth]{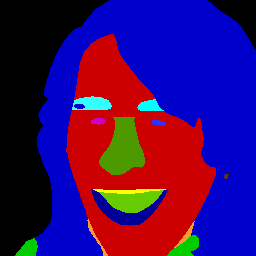} & 
		\includegraphics[width=.13\linewidth]{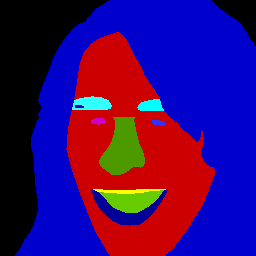} \\[-0.5mm]
		
		\includegraphics[width=.13\linewidth]{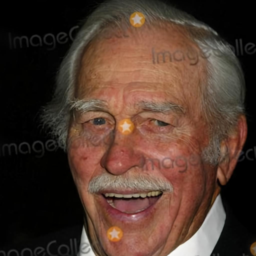} &
		\includegraphics[width=.13\linewidth]{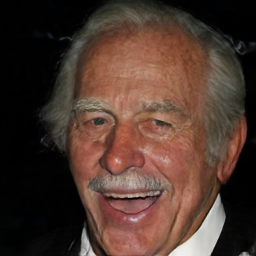} &
		\includegraphics[width=.13\linewidth]{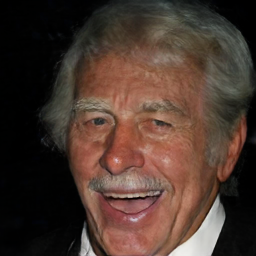} &
		\includegraphics[width=.13\linewidth]{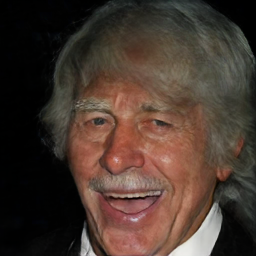} &
		\includegraphics[width=.13\linewidth]{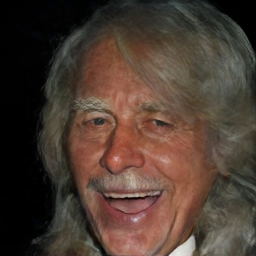} &
		\includegraphics[width=.13\linewidth]{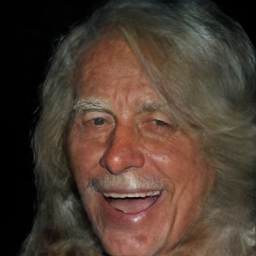} \\[-0.5mm]
		\multicolumn{1}{c!\vrule}{Ground-truth} & \multicolumn{5}{c}{Interpolation to target proportions} \\[-0.5mm]
		\arrayrulecolor{black!50}\midrule
	\end{tabular}
    \vspace{-0.3cm}
	\caption{\small \textbf{Hair manipulation 1: Grow existing hair}. We select subjects with short hair and progressively increase the hair budget. The hair style corresponds to the input ground-truth image-layout pair. Please, zoom in for details.}
	\label{fig:face_edit_1}
	\vspace{-0.3cm}
\end{figure*}

\begin{figure*}
	\setlength\tabcolsep{1.0pt}
	\centering
	\begin{tabular}{cccccc}
		\includegraphics[width=.13\linewidth]{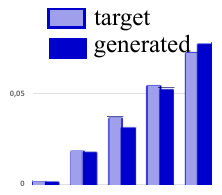} &
		\includegraphics[width=.13\linewidth]{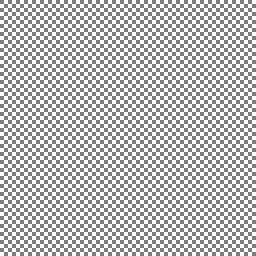} & 
		\includegraphics[width=.13\linewidth]{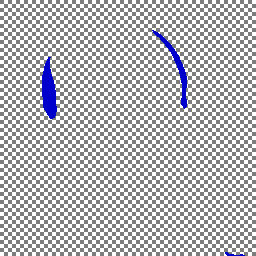} & 
		\includegraphics[width=.13\linewidth]{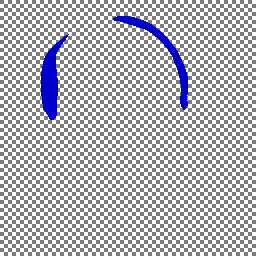} & 
		\includegraphics[width=.13\linewidth]{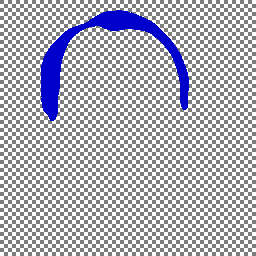} & 
		\includegraphics[width=.13\linewidth]{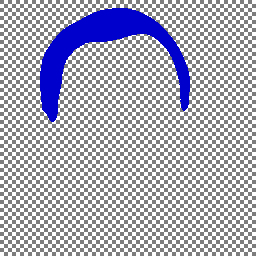} \\[-0.5mm]
		
		\includegraphics[width=.13\linewidth]{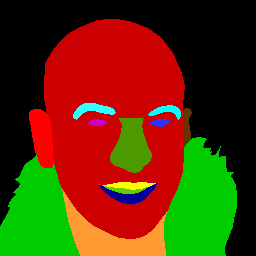} &
		\includegraphics[width=.13\linewidth]{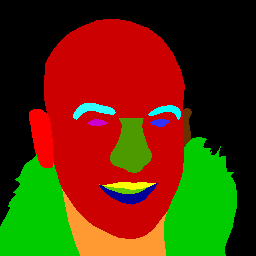} & 
		\includegraphics[width=.13\linewidth]{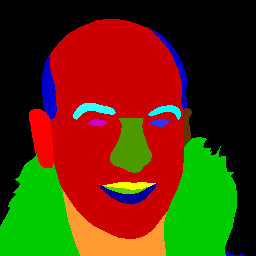} & 
		\includegraphics[width=.13\linewidth]{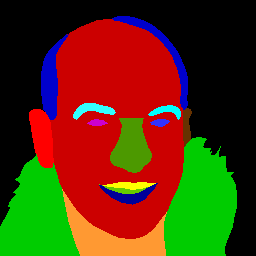} & 
		\includegraphics[width=.13\linewidth]{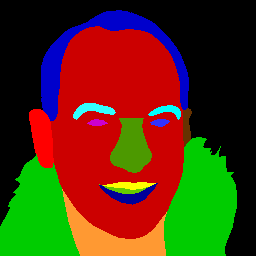} & 
		\includegraphics[width=.13\linewidth]{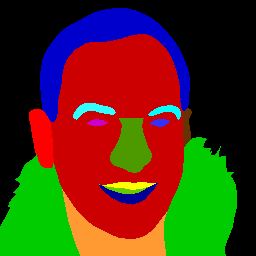} \\[-0.5mm]
		
		\includegraphics[width=.13\linewidth]{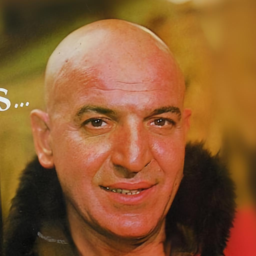} &
		\includegraphics[width=.13\linewidth]{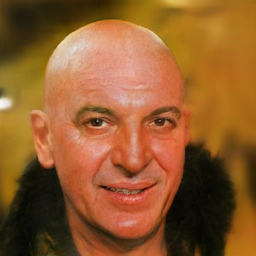} &
		\includegraphics[width=.13\linewidth]{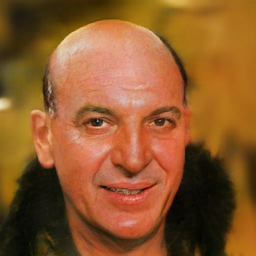} &
		\includegraphics[width=.13\linewidth]{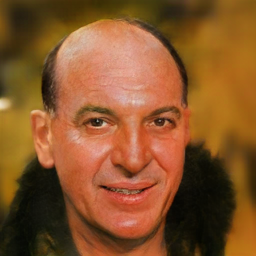} &
		\includegraphics[width=.13\linewidth]{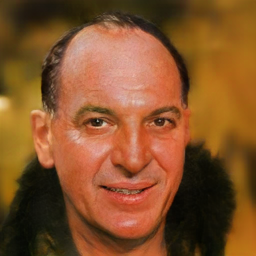} &
		\includegraphics[width=.13\linewidth]{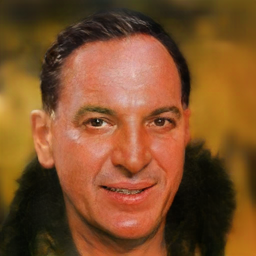} \\ [-0.5mm]
	\end{tabular}
	
	\vspace{1mm}
	
	\begin{tabular}{cccccc}
		\includegraphics[width=.13\linewidth]{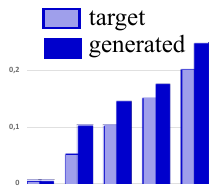} &
		\includegraphics[width=.13\linewidth]{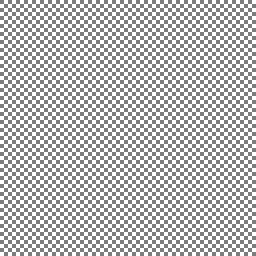} & 
		\includegraphics[width=.13\linewidth]{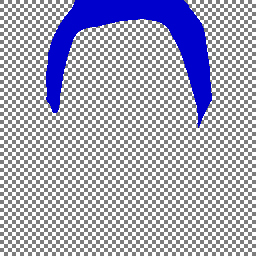} & 
		\includegraphics[width=.13\linewidth]{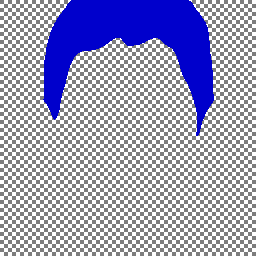} & 
		\includegraphics[width=.13\linewidth]{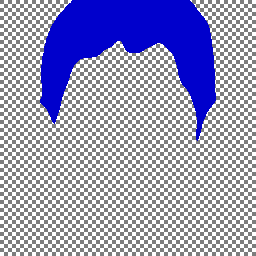} & 
		\includegraphics[width=.13\linewidth]{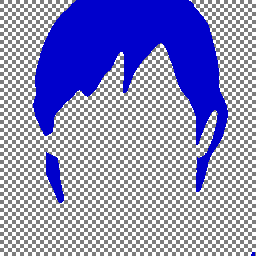} \\[-0.5mm]
		
		\includegraphics[width=.13\linewidth]{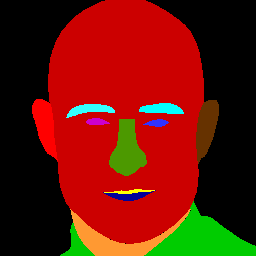} &
		\includegraphics[width=.13\linewidth]{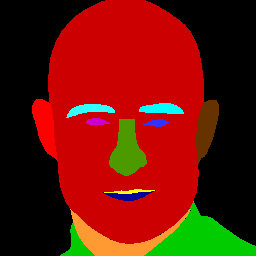} & 
		\includegraphics[width=.13\linewidth]{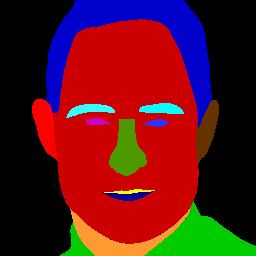} & 
		\includegraphics[width=.13\linewidth]{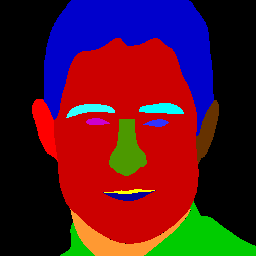} & 
		\includegraphics[width=.13\linewidth]{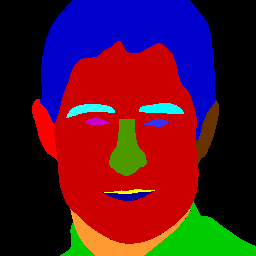} & 
		\includegraphics[width=.13\linewidth]{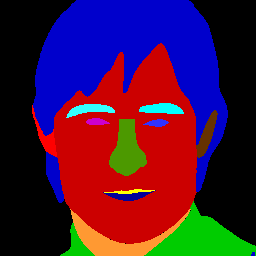} \\[-0.5mm]
		
		\includegraphics[width=.13\linewidth]{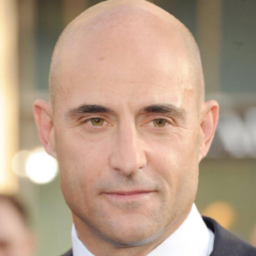} &
		\includegraphics[width=.13\linewidth]{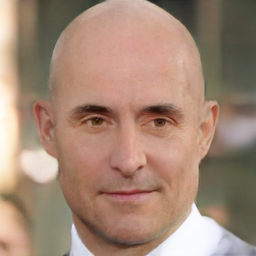} &
		\includegraphics[width=.13\linewidth]{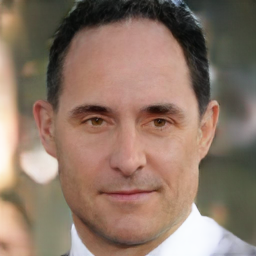} &
		\includegraphics[width=.13\linewidth]{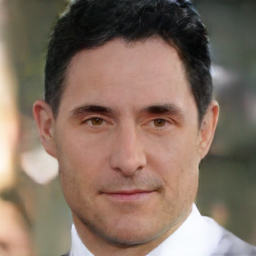} &
		\includegraphics[width=.13\linewidth]{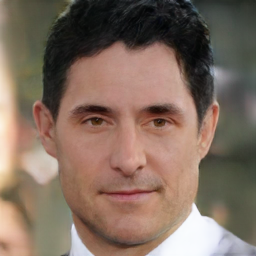} &
		\includegraphics[width=.13\linewidth]{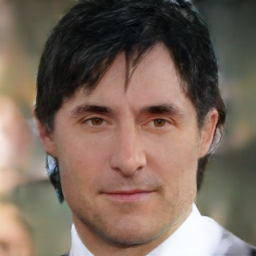} \\ [-0.5mm]
	\end{tabular}
	
	\vspace{1mm}
	
	\begin{tabular}{cccccc}
		\includegraphics[width=.13\linewidth]{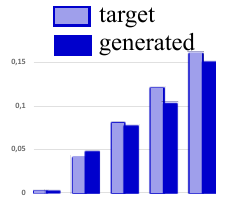} &
		\includegraphics[width=.13\linewidth]{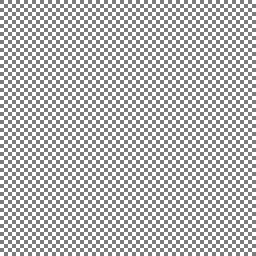} & 
		\includegraphics[width=.13\linewidth]{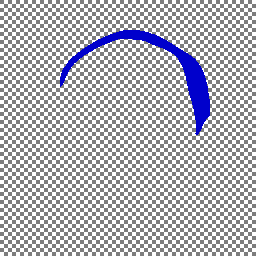} & 
		\includegraphics[width=.13\linewidth]{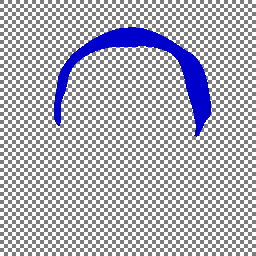} & 
		\includegraphics[width=.13\linewidth]{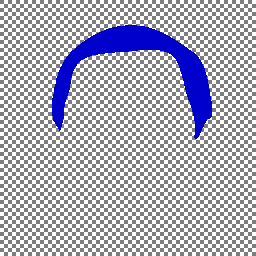} & 
		\includegraphics[width=.13\linewidth]{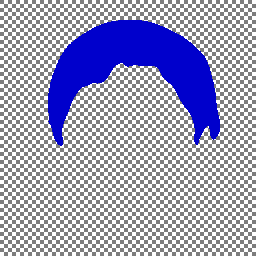} \\[-0.5mm]
		
		\includegraphics[width=.13\linewidth]{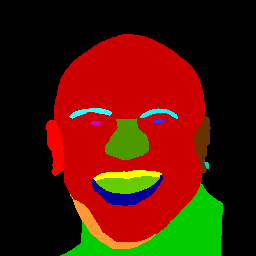} &
		\includegraphics[width=.13\linewidth]{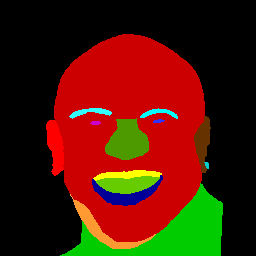} & 
		\includegraphics[width=.13\linewidth]{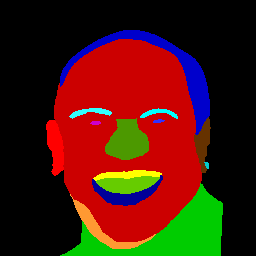} & 
		\includegraphics[width=.13\linewidth]{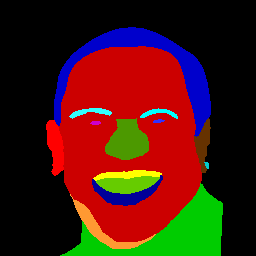} & 
		\includegraphics[width=.13\linewidth]{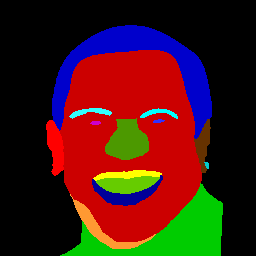} & 
		\includegraphics[width=.13\linewidth]{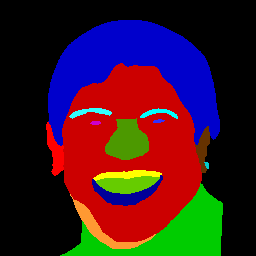} \\[-0.5mm]
		
		\includegraphics[width=.13\linewidth]{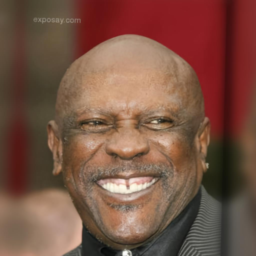} &
		\includegraphics[width=.13\linewidth]{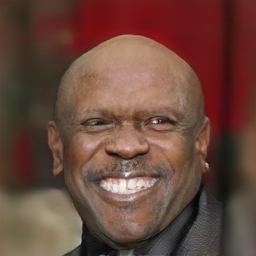} &
		\includegraphics[width=.13\linewidth]{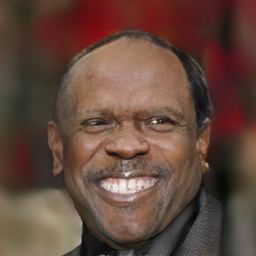} &
		\includegraphics[width=.13\linewidth]{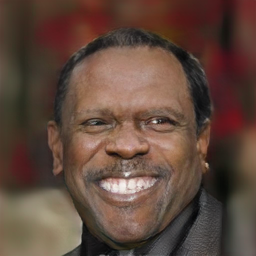} &
		\includegraphics[width=.13\linewidth]{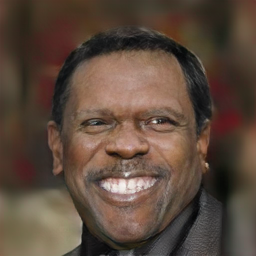} &
		\includegraphics[width=.13\linewidth]{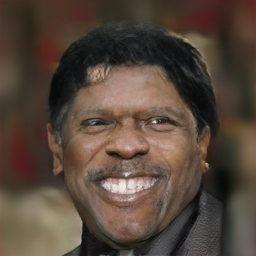} \\[-0.5mm]
		\multicolumn{1}{c!\vrule}{Ground-truth} & \multicolumn{5}{c}{Interpolation to target proportions} \\[-0.5mm]
		\arrayrulecolor{black!50}\midrule
	\end{tabular}
    \vspace{-0.3cm}
	\caption{\small \textbf{Hair manipulation 2: Bald to not bald}. We select bald subjects and progressively increase the hair budget. Since there is no hair initially, the hair style is randomly burrowed from another subject in the training set. Please, zoom in for details.}
	\label{fig:face_edit_2}
	\vspace{-0.3cm}
\end{figure*}

\begin{figure*}
	\setlength\tabcolsep{1.0pt}
	\centering
    \begin{tabular}{cccccc}
		\includegraphics[width=.13\linewidth]{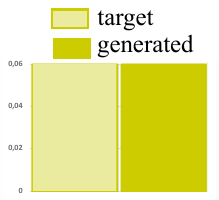} &
		\includegraphics[width=.13\linewidth]{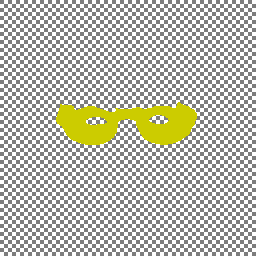} & 
		\includegraphics[width=.13\linewidth]{./figs/suppmat/glasses1/hist.pdf} & 
		\includegraphics[width=.13\linewidth]{./figs/suppmat/glasses1/part.png} & 
		\includegraphics[width=.13\linewidth]{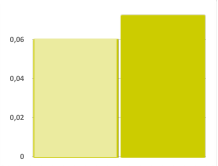} & 
		\includegraphics[width=.13\linewidth]{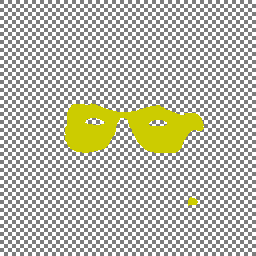} \\[-0.5mm]
		
		\includegraphics[width=.13\linewidth]{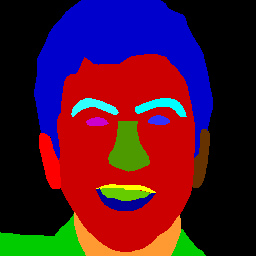} &
		\includegraphics[width=.13\linewidth]{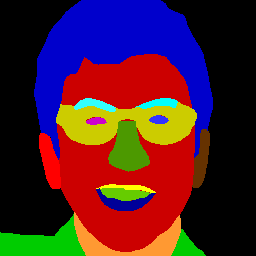} & 
		\includegraphics[width=.13\linewidth]{./figs/suppmat/glasses1/rlay.png} & 
		\includegraphics[width=.13\linewidth]{./figs/suppmat/glasses1/flay.png} & 
		\includegraphics[width=.13\linewidth]{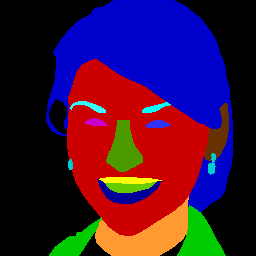} & 
		\includegraphics[width=.13\linewidth]{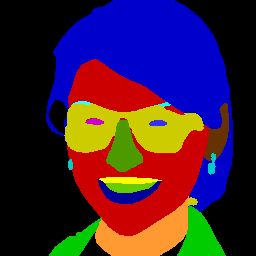} \\[-0.5mm]
		
		\includegraphics[width=.13\linewidth]{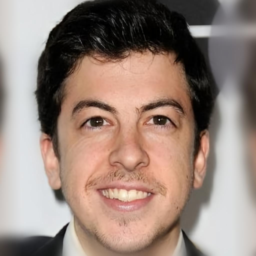} &
		\includegraphics[width=.13\linewidth]{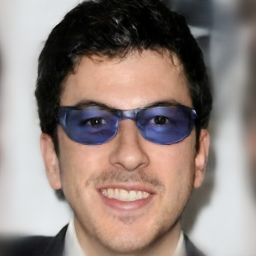} &
		\includegraphics[width=.13\linewidth]{./figs/suppmat/glasses1/rimg.png} &
		\includegraphics[width=.13\linewidth]{./figs/suppmat/glasses1/fimg.png} &
		\includegraphics[width=.13\linewidth]{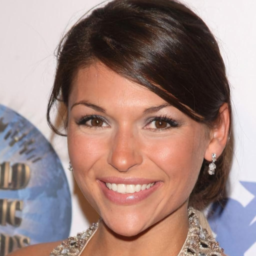} &
		\includegraphics[width=.13\linewidth]{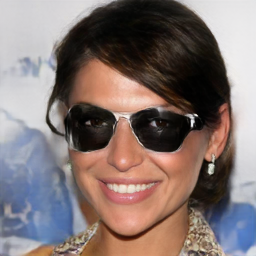} \\[-0.5mm]
	\end{tabular}
	
	\vspace{1mm}
	
	\begin{tabular}{cccccc}
		\includegraphics[width=.13\linewidth]{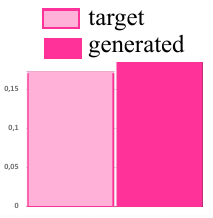} &
		\includegraphics[width=.13\linewidth]{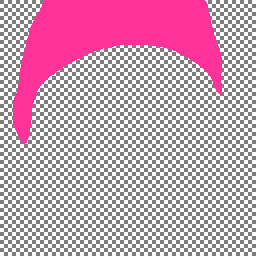} & 
		\includegraphics[width=.13\linewidth]{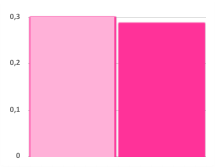} & 
		\includegraphics[width=.13\linewidth]{./figs/suppmat/hat1/part.png} & 
		\includegraphics[width=.13\linewidth]{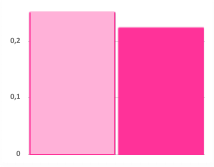} & 
		\includegraphics[width=.13\linewidth]{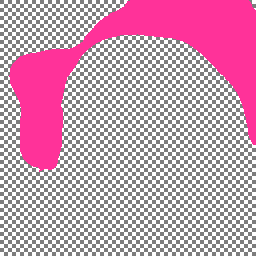} \\[-0.5mm]
		
		\includegraphics[width=.13\linewidth]{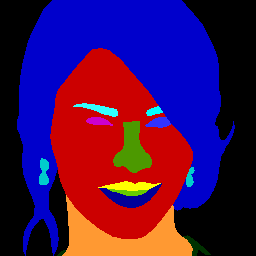} &
		\includegraphics[width=.13\linewidth]{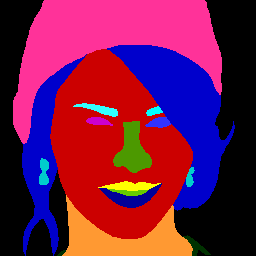} & 
		\includegraphics[width=.13\linewidth]{./figs/suppmat/hat1/rlay.png} & 
		\includegraphics[width=.13\linewidth]{./figs/suppmat/hat1/flay.png} & 
		\includegraphics[width=.13\linewidth]{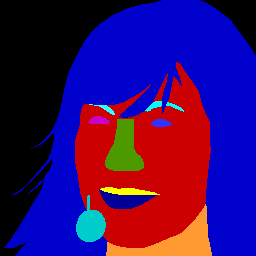} & 
		\includegraphics[width=.13\linewidth]{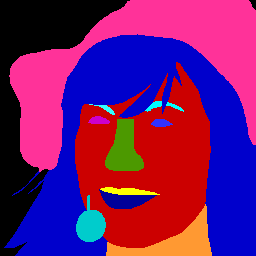} \\[-0.5mm]
		
		\includegraphics[width=.13\linewidth]{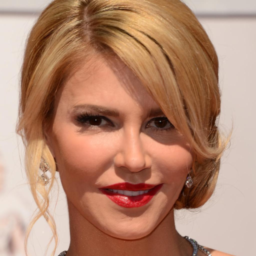} &
		\includegraphics[width=.13\linewidth]{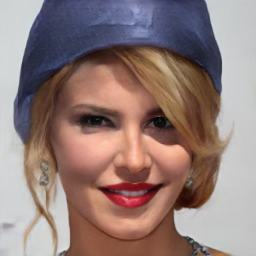} &
		\includegraphics[width=.13\linewidth]{./figs/suppmat/hat1/rimg.png} &
		\includegraphics[width=.13\linewidth]{./figs/suppmat/hat1/fimg.png} &
		\includegraphics[width=.13\linewidth]{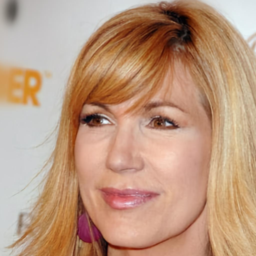} &
		\includegraphics[width=.13\linewidth]{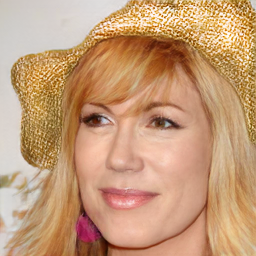} \\[-0.5mm]
	\end{tabular}
	
	\vspace{1mm}
	
	\begin{tabular}{cccccc}
		\includegraphics[width=.13\linewidth]{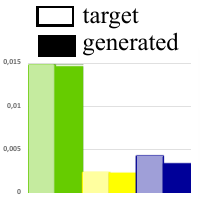} &
		\includegraphics[width=.13\linewidth]{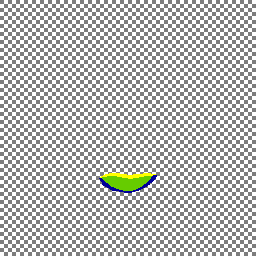} & 
		\includegraphics[width=.13\linewidth]{./figs/suppmat/teeth1/hist.pdf} & 
		\includegraphics[width=.13\linewidth]{./figs/suppmat/teeth1/part.png} & 
		\includegraphics[width=.13\linewidth]{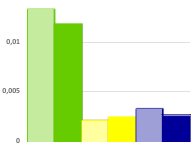} & 
		\includegraphics[width=.13\linewidth]{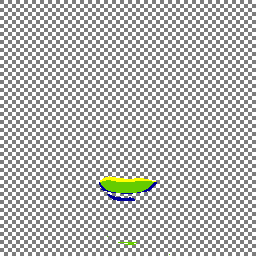} \\[-0.5mm]
		
		\includegraphics[width=.13\linewidth]{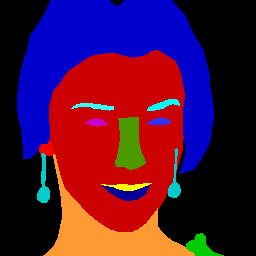} &
		\includegraphics[width=.13\linewidth]{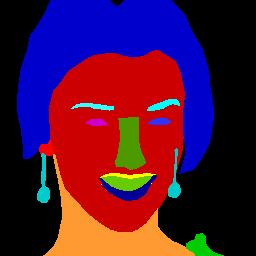} & 
		\includegraphics[width=.13\linewidth]{./figs/suppmat/teeth1/rlay.png} & 
		\includegraphics[width=.13\linewidth]{./figs/suppmat/teeth1/flay.png} & 
		\includegraphics[width=.13\linewidth]{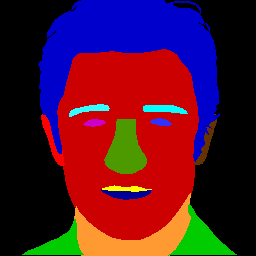} & 
		\includegraphics[width=.13\linewidth]{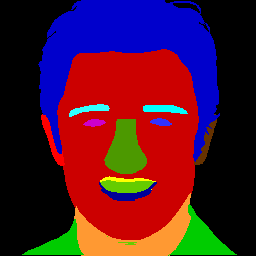} \\[-0.5mm]
		
		\includegraphics[width=.13\linewidth]{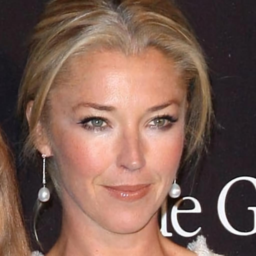} &
		\includegraphics[width=.13\linewidth]{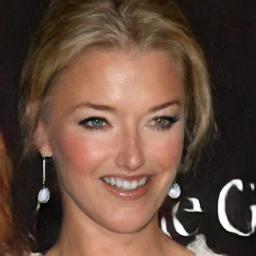} &
		\includegraphics[width=.13\linewidth]{./figs/suppmat/teeth1/rimg.png} &
		\includegraphics[width=.13\linewidth]{./figs/suppmat/teeth1/fimg.png} &
		\includegraphics[width=.13\linewidth]{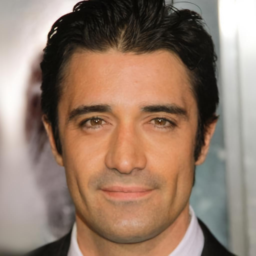} &
		\includegraphics[width=.13\linewidth]{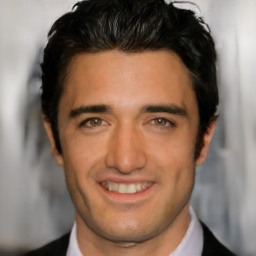} \\[-0.5mm]
		Ground-truth & \multicolumn{1}{c!\vrule}{Generated} & Ground-truth & \multicolumn{1}{c!\vrule}{Generated} & Ground-truth & Generated\\[-0.5mm]
		\arrayrulecolor{black!50}\midrule
	\end{tabular}
    \vspace{-0.3cm}
	\caption{\small \textbf{Manipulation of diverse semantic attributes.} Glasses (1\textsuperscript{st} row), hat (2\textsuperscript{nd}), teeth (3\textsuperscript{rd}). Please, zoom in for details.}
	\label{fig:face_edit_3}
	\vspace{-0.3cm}
\end{figure*}

We provide additional qualitative results of scene generation in Figures~\ref{fig:idd} and~\ref{fig:city}, and of face editing in Figures~\ref{fig:face_edit_1},~\ref{fig:face_edit_2} and~\ref{fig:face_edit_3}.
These figures are best viewed in color.

\end{document}